%% file: main.tex
\pdfoutput=1

\documentclass[11pt]{article}

\usepackage[preprint]{acl}

\input{macro}

\title{Do Vision-Language Models Have Internal World Models? \\ Towards an Atomic Evaluation}

\input{authors}

\begin{document}

\input{teaser}
\maketitle

\begin{abstract}
Internal world models (WMs) enable agents to understand the world's state and predict transitions, serving as the basis for advanced deliberative reasoning.
Recent large Vision-Language Models (VLMs), such as OpenAI o3, GPT-4o and Gemini, exhibit potential as general-purpose WMs. While the latest studies have evaluated and shown limitations in specific capabilities such as visual understanding, a systematic evaluation of VLMs' fundamental WM abilities remains absent. Drawing on comparative psychology and cognitive science, we propose a two-stage framework that assesses {\it perception} (visual, spatial, temporal, quantitative, and motion) and {\it prediction} (mechanistic simulation, transitive inference, compositional inference) to provide an atomic evaluation of VLMs as WMs. Guided by this framework, we introduce \benchmark, a large-scale benchmark comprising 23 fine-grained evaluation dimensions across 6 diverse simulated environments with controlled counterfactual simulations. Through 660 experiments on 15 latest commercial and open-source VLMs, we find that these models exhibit striking limitations in basic world modeling abilities. For instance, almost all models perform at near-random accuracy when distinguishing motion trajectories. Additionally, they lack disentangled understanding---e.g., some models tend to believe blue objects move faster than green ones. More rich results and analyses reveal significant gaps between VLMs and human-level world modeling.  

\end{abstract}


\input{MainSections/01-intro}

\input{MainSections/02-Conceptual_Framework}

\input{MainSections/03-Methodology}

\input{MainSections/04-Experiments}

\input{MainSections/05-Discussion}

\input{MainSections/06-Related_work}

\input{MainSections/07-Conclusion}

\input{MainSections/08-Limitation}

\bibliography{2024,custom, custom_wm}

\clearpage
\appendix
\section*{\Large Appendix}

\input{Appendices/a1_framework}
\input{Appendices/a2_benchmark}
\input{Appendices/a3_simulatoir}
\input{Appendices/a4_results}
\input{Appendices/a5_reproduce}

\end{document}

%% file: macro.tex
\usepackage{times}
\usepackage{latexsym}

\usepackage[T1]{fontenc}

\usepackage[utf8]{inputenc}

\usepackage{microtype}

\usepackage{inconsolata}

\usepackage{graphicx}
\usepackage{hyperref}       
\usepackage{url}            
\usepackage{xurl}
\usepackage{booktabs}       
\usepackage{amsfonts}       
\usepackage{amsmath}
\usepackage{nicefrac}       
\usepackage{microtype}      
\usepackage{tabularx}
\usepackage{booktabs}
\usepackage{multirow} 
\usepackage{array}
\usepackage{arydshln}
\usepackage{color, colortbl}
\usepackage{subcaption}
\usepackage{tabu}
\usepackage{pifont}
\usepackage{makecell}
\usepackage{tcolorbox} 
\usepackage{lipsum}  
\usepackage{paralist}
\usepackage{threeparttable}
\usepackage{hhline}
\usepackage{scalerel,xparse}
\usepackage{appendix}
\usepackage{float}
\usepackage{longtable}
\usepackage{colortbl}
\usepackage{pgfplotstable}
\pgfplotsset{compat=1.17}

\usepackage{xspace}

\usepackage[shortlabels]{enumitem}
\setlist[itemize]{leftmargin=*}
\setlist[enumerate]{leftmargin=*}
\usepackage[normalem]{ulem}
\newcommand{\benchmark}{\texttt{WM-ABench}\xspace}

\NewDocumentCommand{\bert}{ mO{} }{\textcolor{ForestGreen}{\textsuperscript{\textit{Qiyue}}\textsf{\textbf{\small[#1]}}}}

\usepackage{todonotes}

\definecolor{lightblue}{HTML}{C7E0EB} 
\definecolor{deepblue}{HTML}{82B9D3}  
\definecolor{midblue}{HTML}{7E9CD2}   

\definecolor{lightred}{HTML}{F0C8C6}  
\definecolor{midred}{HTML}{EAB0AB}    
\definecolor{darkred}{HTML}{DB776F}   

%% file: authors.tex
\author{%
Qiyue Gao$^{*}$, Xinyu Pi$^{*}$, Kevin Liu, Junrong Chen, Ruolan Yang\\
\textbf{Xinqi Huang, Xinyu Fang, Lu Sun, Gautham Kishore, Bo Ai, Stone Tao, Mengyang Liu}\\
\textbf{Jiaxi Yang, Chao-Jung Lai, Chuanyang Jin, Jiannan Xiang, Benhao Huang, Zeming Chen}\\
\textbf{David Danks, Hao Su, Tianmin Shu, Ziqiao Ma, Lianhui Qin, Zhiting Hu}\\
Maitrix.org\quad UC San Diego\quad JHU\quad Cornell Tech\quad EPFL\quad UMich \\
\url{https://wm-abench.maitrix.org/}\qquad *\textit{equal contribution}
}

%% file: MainSections/01-intro.tex
\section{Introduction}  

World models (WMs) in agents provide internal representations of the external world \citep{johnson1983mental,wooldridge1995intelligent}.
They simulate how the current state transforms to the next \citep{kappes2016mental, russell2016artificial}, enabling agents to perform deliberative planning by predicting probable future states and choosing the most favorable one \citep{ha2018world, lecun2022path, hu2023language, ai2024robopack, tian2025diffusion}. 
Different environments operate under distinct mechanisms and dynamics---for example, the transition mechanics of autonomous vehicles, surgical robots, and rover-based spacecraft vary significantly. 
Prior work handled this diversity by building environment-specific world models, but they failed to adapt across various real applications.

Recent large-scale Vision-Language Models \citep[VLMs;][\textit{inter alia}]{Gemini2023,openai2024gpt4, anthropic2024claude,li2024llava,OpenAI2025o3o4minisystemcard} enhance generalist LLMs with visual semantics and encapsulate extensive knowledge of world dynamics, making them promising potential world models for general domains. 
Unlike video generative world models which directly generate the next states, VLMs can reason in their latent representation space and forecast via language.
However, their language grounding and world simulation capabilities may still be insufficient in various aspects.
For example, existing benchmarks have revealed their vulnerability in visual or spatiotemporal perception~\citep{goyal2020rel3dminimallycontrastivebenchmark,shangguan2024tomatoassessingvisualtemporal,fu2024blink,zhang2025do} and future state prediction driven by intuitive physics~\citep{bakhtin2019phyrenewbenchmarkphysical,yi2020clevrer,bear2022physion}.
These limitations underscore the need for a more systematic evaluation.
A robust world model must integrate multiple fundamental abilities in perception and prediction, yet previous studies that assess these aspects in isolation provide only a partial view. To address this, we propose an {\it atomic} evaluation framework, systematically testing the essential aspects (and their interactions) of a VLM's internal world model from first principles.

With theories and evidence from comparative psychology and cognitive science \citep{spelke2000core, knill2004bayesian, olmstead2015comparative}, we first present a systematic conceptual framework to formalize the functioning of world models (Figure~\ref{fig:framework}).
We decompose the process into two stages: (1) \textit{perception} stage, which involves \textit{visual}, \textit{spatial}, \textit{temporal}, \textit{quantitative}, and \textit{motion} perceptions \citep{Baillargeon1985-uu,merleau2004world, coren2004sensation,hoffmann2011ontogeny}; and (2) \textit{prediction} stage, which involves \textit{mechanistic simulation}, \textit{transitive inference}, and \textit{compositional inference} \citep{hegarty2004mechanical, barsalou2008grounded,prystawski2023thinkstepstepreasoning}.

Following this framework, we create \benchmark, the World Model Atomic Benchmark that covers 23 fine-grained dimensions (Figure~\ref{fig:alltasks}) of world modeling and over 100,000 instances curated from 6 different simulators~\citep{Dosovitskiy17,threedworld,szot2021habitat2,taomaniskill3,bear2022physion,gu2023maniskill2}. 
By systematically manipulating environmental factors and simulating counterfactual actions, we generate incorrect states for models to differentiate from the correct ones to ensure {\it controlled} studies. 
We also measure human performance to verify the fairness and solvability of our problems.

We conduct 660 experiments on 15 state-of-the-art VLMs and find that, while they excel in certain aspects of visual and quantitative perception, they are surprisingly limited in many other dimensions (Figure~\ref{fig:framework}).

Specifically, VLMs exhibit
(1) weak perception of space, time, and motion;
(2) insufficient knowledge of physical causality in intuitive physics and agentic actions;
(3) limited transitive and compositional reasoning capabilities, showing near-random accuracy.
Our further analyses reveal a lack of independent and robust world representations in VLMs, e.g., mistakenly associating color with speed.
These findings reveal significant gaps between current VLMs and human-level world modeling, shedding light on the need for deeper understanding, grounding, and reasoning over the perceptual world and its transition mechanisms before VLMs can truly serve as generalist world models.

%% file: MainSections/02-Conceptual_Framework.tex
\vspace{-3pt}
\section{The Dual-Stage Conceptual Framework}
\vspace{-2pt}

We view world modeling as a two-stage process of \textit{perception} and \textit{prediction} \citep{knill2004bayesian, ha2018world}. 
In the first stage, agents form internal representations of the current state by sensing and encoding environmental stimuli.
In the second stage, agents use these internal representations to extrapolate future states, refining their model whenever new ground-truth observations arrive.
This dual-stage framework explains (1) how raw sensory signals are converted into compact world representations; and (2) how these representations then guide forward simulations.
The two stages of WMs are core functions in agents' advanced planning and decision-making.
We brief our formulation here and leave the in-depth discussion in Appendix \ref{conceptualframework}.

\vspace{-5pt}
\subsection{Perception Stage}
Perception involves extracting and organizing essential information from multi-sensory cues \citep{merleau2004world, coren2004sensation}.
It is not just bottom-up signal processing but also top-down inference grounded in prior knowledge. 
For instance, many intelligent animals (e.g., crows, dolphins, chimpanzees) grasp \textit{object permanence} by maintaining accurate representations of hidden or partially occluded objects \citep{Baillargeon1985-uu, hoffmann2011ontogeny}. 
While real-world perception spans multitudinous modalities (time, vision, temperature, humidity, audition, proprioception, etc.), we only focus on a limited number of perceptual dimensions. 
From an epistemic perspective, we design these dimensions to maximize the information captured about the external world while minimizing representational dimensions, ensuring they remain mutually orthogonal. 
From the practical perspective, we only cover the dimensions that modern VLMs can access.
In this work, we consider 5 perceptual dimensions in our framework: space, time, motion, quantity, and vision. 
To make models' perceptual competency empirically testable along each dimension, we further break the major 5 perceptual dimensions into sub-dimensions:

\vspace*{-5pt}
\paragraph{Space and Time.} 
All entities in spacetime must be located at some position and occupy some room or period, i.e., \textit{extension}. 
Any pair of entities to exist in spacetime necessarily have spatiotemporal relations (e.g., front, left, before, after). 
Thus, we break spacetime down into \textit{position}, \textit{extension}, and \textit{relations}
\citep{reichenbach2012philosophy}.

\vspace*{-5pt}
\paragraph{Motion.} 
At every moment, motion can be described as a vector with speed and direction. 
The integration of motion along time forms a trajectory. 
Thus we consider \textit{direction}, \textit{speed}, and \textit{trajectory} \citep{johansson1975visual}.

\vspace*{-5pt}
\paragraph{Quantity.} 
Axiomatically, quantities can either be discrete or continuous. 
For any pair of quantities, there could be quantitative relations (e.g., more, less, $n$-times) \citep{kaufman1949discrimination,hurst2024continuous}. 
We consider \textit{discrete}, \textit{continuous}, and \textit{relations} \cite{kadosh2015oxford} quantities in this study.

\vspace*{-5pt}
\paragraph{Vision.} 
In contrast with the previous four dimensions, the vision channel is extremely broad and hard to enumerate, including \textit{orientation}, \textit{density}, \textit{edge}, \textit{color}, \textit{shape}, to name a few \cite{marr2010vision}. 
We consider salient features like \textit{color}, \textit{shape}, and \textit{material} (i.e., texture and reflexivity).

\vspace{-5pt}
\subsection{Prediction Stage}
Once current-state representations are established, the agent must predict how future states evolve in response to both natural dynamics and possible actions. 
We distinguish 3 primary sub-dimensions: 

\vspace*{-5pt}
\paragraph{Mechanistic Simulation.}
Agents should understand the causality of intuitive physical dynamics (e.g., motion, collisions) and intentional actions to simulate the next state \citep{hegarty2004mechanical, barsalou2008grounded}. 
For example, predicting how one ball bounces off a wall or another ball depends on basic principles like momentum or elasticity.

\vspace*{-5pt}
\paragraph{Transitive Inference.}
Multi-step forecasts are often needed for tasks requiring long-horizon planning \citep{prystawski2023thinkstepstepreasoning}. 
Rather than only predicting the immediate next state, robust world models should extrapolate further into the future by chaining intermediate predictions.

\vspace*{-5pt}
\paragraph{Compositional Inference.}
Real-world scenarios usually involve multiple interacting objects and agents (e.g., two incoming balls hitting a third one from different directions). 
Agents must merge known mechanisms to predict novel outcomes, even if that specific combination has not been observed~\cite{XU200997, Eckert2021TheAL, doi:10.1073/pnas.1003095107}.
This requires compositional reasoning, where partial pre-conditions (e.g., ``hit from the left'' plus ``hit from the right'') merge into an overall post-condition (e.g., ``moves straight up'').

\begin{figure*}[ht!]
    \centering
    \includegraphics[width=1.0\linewidth]{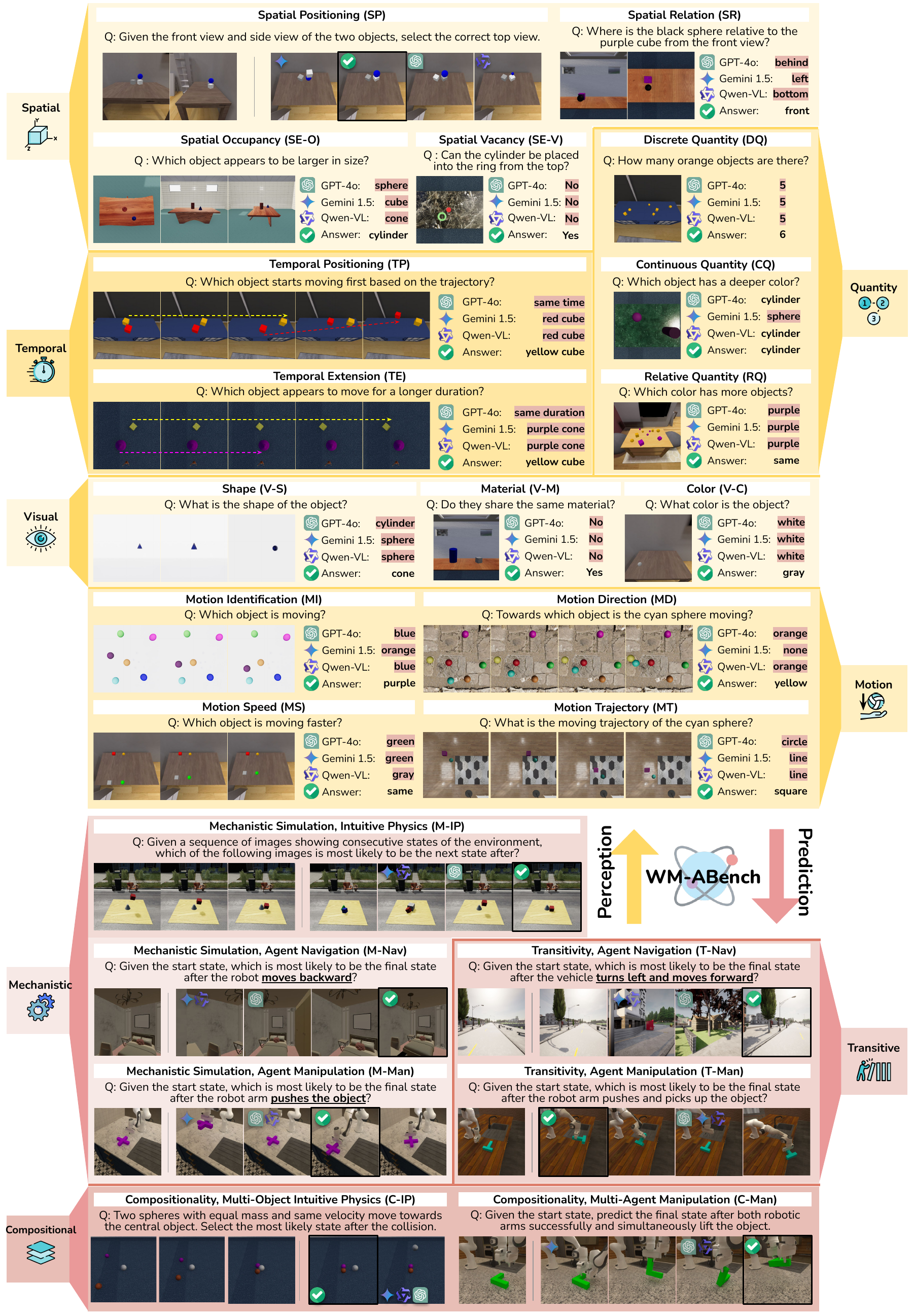}
    \vspace{-20pt}
    \caption{Overview of \benchmark{} tasks. The Perception stage (top) covers Spatial, Temporal, Visual, Quantity, and Motion dimensions, each shown with example questions and outputs. The Prediction stage (bottom) includes Mechanistic Simulation, which covers Intuitive Physics (e.g., drop), Agent Navigation (e.g., turn left), and Agent Manipulation (e.g., push), plus Transitivity and Compositionality tasks that build on these transitions.}
    \vspace{-15pt}
    \label{fig:alltasks}
\end{figure*}

%% file: MainSections/03-Methodology.tex
\vspace{-3pt}
\section{The \benchmark Benchmark}
\vspace{-2pt}

Following the above taxonomy, we design and implement corresponding tasks for benchmarking VLMs as WMs.
Figure \ref{fig:alltasks} presents a visual overview of our complete benchmark tasks, offering readers a high-level understanding of the framework.

\input{tables/perception_stv_results}
\input{tables/perception_mq_results}

\vspace*{-5pt}
\paragraph{Controlled Experiment and Causal Analysis.} 
To ensure controlled evaluation, we exhaustively iterate over all dimensions, keep all other dimensions fixed as independent variables, and allow only one to vary at each data point.
This methodology, which holds all variables constant except the independent ones across data points, allows us to draw causal conclusions (e.g., changing color causes the model to misperceive size).
Rather than claiming generality or complete coverage~\citep{raji2021ai,saxon2024benchmarks}, this benchmark aims to provide a precise, atomic diagnosis of models' perception and prediction, establishing a clear checklist for VLMs as world models.

\vspace*{-5pt}
\paragraph{Fighting Shortcuts and Spurious Correlations.} 
Pre-trained models such as VLMs are known to rely on shortcuts and spurious correlations \citep{ye2024mmspubench,steinmann2024navigating}.
To test whether VLMs can truly simulate and extrapolate into the next states, rather than relying on some spurious correlations, we generate hard negative options in our benchmark. 
We consider 2 methodologies to generate counterfactual states: counterfactual action and counterfactual previous states.  
Consider a ground truth transition triplet $(S_t^*, a^*, S_{t+1}^*)$.
For counterfactual action-based option generation, we fix the ground-truth previous state and perturb the action, and the transition becomes $(S_{t}^{*}, a', S_{t+1}')$. 
For counterfactual state-based option generation, we keep the ground-truth action, and perturb the previous state, so that the transition becomes $(S_{t}', a^{*}, S_{t+1}'')$.
False options generated under these methodologies often exhibit high visual similarity to the ground truth state, requiring models to possess a genuine understanding of world dynamics to distinguish and eliminate counterfactual states.

\input{tables/prediction_results_full}

\vspace*{-5pt}
\paragraph{Data Collection.}

To rigorously evaluate models, a large and diverse dataset of test cases is essential. While manual data collection is possible, it is costly and often impractical for obtaining images where only a single factor varies for controlled studies. 
Therefore, we utilize compute-scalable simulation frameworks to synthetically generate a substantial number of test cases, minimizing human labor while ensuring precise control over variations.
To avoid bias towards one single environment, we use a wide variety of simulation frameworks to create data, including ThreeDWorld \citep[TDW;][]{threedworld}, ManiSkill \citep{taomaniskill3,gu2023maniskill2}, Habitat 2.0 \citep{szot2021habitat2}, Physion \citep{bear2022physion}, and Carla \citep{Dosovitskiy17}. 
Our diverse set of simulators allows the benchmark to incorporate various world dynamics that align with human intuition while leveraging a broad range of assets to address diverse questions. 

%% file: tables/perception_stv_results.tex
\begin{table*}[t]
\begin{minipage}{0.28\textwidth}
\centering
\small
\includegraphics[width=1.0\textwidth]{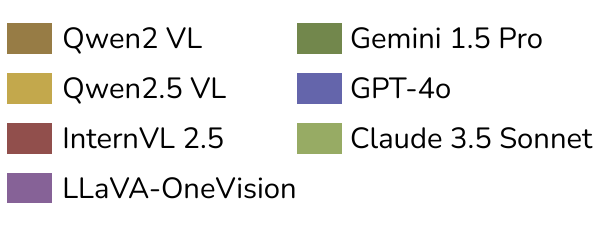}
\includegraphics[width=1.0\textwidth]{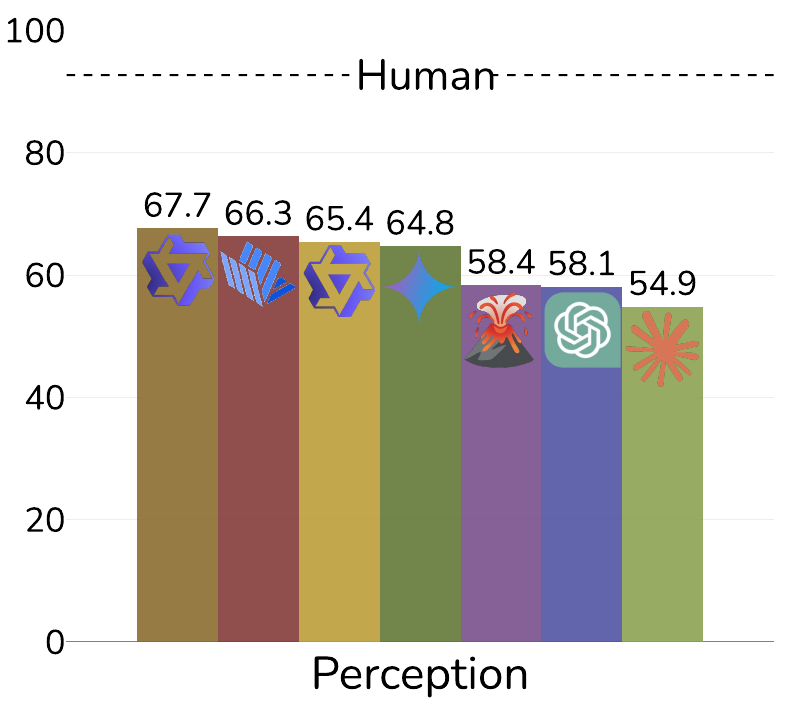}
\end{minipage}
\hspace{-8pt}
\begin{minipage}{0.57\textwidth}
\scalebox{0.685}{
\begingroup
\renewcommand{\arraystretch}{0.8}
\setlength{\tabcolsep}{1.5pt}
\begin{tabular}{l
cc cc cc cc
cc cc
}
\toprule
& \multicolumn{8}{c}{\textbf{Space}} 
& \multicolumn{4}{c}{\textbf{Time}} \\
\cmidrule(lr){2-9}\cmidrule(lr){10-13}
\textbf{Model} 
& \multicolumn{2}{c}{\textbf{SR}} 
& \multicolumn{2}{c}{\textbf{SE-V}} 
& \multicolumn{2}{c}{\textbf{SE-O}} 
& \multicolumn{2}{c}{\textbf{SP}}
& \multicolumn{2}{c}{\textbf{TP}} 
& \multicolumn{2}{c}{\textbf{TE}} \\
\cmidrule(lr){2-3}\cmidrule(lr){4-5}\cmidrule(lr){6-7}
\cmidrule(lr){8-9}\cmidrule(lr){10-11}\cmidrule(lr){12-13}
& \textbf{Mani.} & \textbf{TDW} 
& \textbf{Mani.} & \textbf{TDW} 
& \textbf{Mani.} & \textbf{TDW} 
& \textbf{Mani.} & \textbf{TDW}
& \textbf{Mani.} & \textbf{TDW}
& \textbf{Mani.} & \textbf{TDW}\\

\midrule
\rowcolor[HTML]{E6ECF2}
\textit{Open-source Models} & & & & & & & & & & & & \\

NVILA 
& \cellcolor{lightblue}53.7  
& \cellcolor{lightblue}40.4  

& \cellcolor{lightblue}55.0  
& \cellcolor{lightred}74.4  

& \cellcolor{lightred}84.9  
& \cellcolor{darkred}92.9  

& \cellcolor{midblue}35.4  
& \cellcolor{midblue}26.9  

& \cellcolor{midblue}35.5  
& \cellcolor{lightblue}59.1  

& \cellcolor{midblue}32.7  
& \cellcolor{midblue}9.0  
\\

QWen2-VL-72b 
& \cellcolor{lightblue}42.2
& \cellcolor{midred}72.6

& \cellcolor{midblue}42.6
& \cellcolor{lightblue}57.0

& \cellcolor{midred}87.6
& \cellcolor{darkred}93.6

&  \cellcolor{lightred}58.9
& \cellcolor{lightblue}47.1

& \cellcolor{lightblue}46.5
& \cellcolor{midred}75.5

& \cellcolor{midblue}36.4
& \cellcolor{lightblue}53.4
\\

QWen2.5-VL-72b 
& \cellcolor{lightred}57.8
& \cellcolor{midred}70.6

& \cellcolor{midblue}49.5
& \cellcolor{lightblue}57.7

& \cellcolor{darkred}97.5
& \cellcolor{darkred}93.5

& \cellcolor{midblue}36.1
& \cellcolor{lightblue}45.5

& \cellcolor{lightblue}48.4
& \cellcolor{lightblue}52.9

& \cellcolor{midblue}35.0
& \cellcolor{lightblue}53.5
\\

InternVL2.5-78b
& \cellcolor{lightblue}44.2
& \cellcolor{lightred}65.8

& \cellcolor{midblue}51.3
& \cellcolor{lightred}73.5

& \cellcolor{darkred}98.4
& \cellcolor{darkred}93.0

& \cellcolor{midblue}31.3
& \cellcolor{midblue}30.9

& \cellcolor{midblue}43.5
& \cellcolor{midred}63.6

& \cellcolor{midblue}43.9
& \cellcolor{lightred}59.7
\\

Llama 3.2 vision-90b
& \cellcolor{lightblue}50.7  
& \cellcolor{lightblue}47.5  

& \cellcolor{midblue}50.0  
& \cellcolor{midblue}43.8  

& \cellcolor{lightred}78.6  
& \cellcolor{midblue}25.2  

& \cellcolor{midblue}25.0  
& \cellcolor{midblue}24.6  

& \cellcolor{midblue}33.4  
& \cellcolor{midblue}16.8  

& \cellcolor{midblue}38.0  
& \cellcolor{midblue}37.2  
\\  

LLaVA-OneVision
& \cellcolor{midred}71.0
& \cellcolor{lightred}61.8

& \cellcolor{lightblue}56.6
& \cellcolor{midblue}53.1

& \cellcolor{darkred}96.5
& \cellcolor{darkred}90.9

& \cellcolor{midblue}24.8
& \cellcolor{midblue}22.7

& \cellcolor{midblue}38.5
& \cellcolor{midblue}31.1

& \cellcolor{midblue}32.2
& \cellcolor{midblue}25.8
\\

\midrule
\rowcolor[HTML]{E6ECF2}
\textit{Closed-source Models} & & & & & & & & & & & & \\


GPT-4o-mini  
& \cellcolor{midblue}37.6  
& \cellcolor{midblue}37.7  

& \cellcolor{lightblue}61.3  
& \cellcolor{lightred}69.1  

& \cellcolor{darkred}98.2  
& \cellcolor{midblue}35.3  

& \cellcolor{midblue}37.0
& \cellcolor{midblue}38.1  

& \cellcolor{midblue}31.7  
& \cellcolor{lightblue}47.1  

& \cellcolor{midblue}29.0  
& \cellcolor{midblue}40.6  
\\ 

GPT-4o
& \cellcolor{lightblue}55.1
& \cellcolor{lightred}58.7

& \cellcolor{lightred}66.1
& \cellcolor{lightred}70.4

& \cellcolor{darkred}96.6
& \cellcolor{lightred}78.9

& \cellcolor{lightblue}47.4
& \cellcolor{lightblue}47.5

& \cellcolor{midblue}37.5
& \cellcolor{lightblue}45.7

& \cellcolor{midblue}35.0
& \cellcolor{midblue}25.0
\\

Claude 3.5 Sonnet
& \cellcolor{lightblue}54.6
& \cellcolor{lightred}59.9

& \cellcolor{lightblue}64.2
& \cellcolor{lightblue}58.0

& \cellcolor{lightred}73.2
& \cellcolor{lightred}72.6

& \cellcolor{lightblue}41.7
& \cellcolor{midblue}31.9

& \cellcolor{midblue}42.0
& \cellcolor{lightblue}49.1

& \cellcolor{midblue}32.2
& \cellcolor{midblue}43.7
\\

Gemini-1.5-flash 
& \cellcolor{lightblue}54.5  
& \cellcolor{lightred}59.1  

& \cellcolor{midblue}50.0  
& \cellcolor{midblue}50.0  

& \cellcolor{midred}80.6  
& \cellcolor{lightred}79.6  

& \cellcolor{midblue}34.0  
& \cellcolor{midblue}21.8  

& \cellcolor{midblue}37.5  
& \cellcolor{lightblue}50.9  

& \cellcolor{midblue}33.4  
& \cellcolor{lightred}56.1  
\\

Gemini-1.5-pro
& \cellcolor{lightblue}49.1  
& \cellcolor{lightred}65.0

& \cellcolor{lightblue}64.0
& \cellcolor{lightred}68.5

& \cellcolor{midred}83.6
& \cellcolor{darkred}92.9

& \cellcolor{lightblue}44.7
& \cellcolor{lightblue}43.4

& \cellcolor{midblue}42.8
& \cellcolor{lightred}60.1

& \cellcolor{midred}77.0
& \cellcolor{midblue}42.2 
\\

GPT-4.5*
& \cellcolor{midred}72.0
& \cellcolor{midred}79.0
& \cellcolor{lightblue}55.0
& \cellcolor{lightblue}52.0
& \cellcolor{darkred}95.0
& \cellcolor{darkred}92.0
& \cellcolor{lightred}57.0
& \cellcolor{midblue}38.0
& \cellcolor{lightblue}47.0
& \cellcolor{midred}66.0
& \cellcolor{lightred}55.0
& \cellcolor{lightred}67.0 \\

OpenAI o3*
& \cellcolor{darkred}88.0
& \cellcolor{darkred}88.0
& \cellcolor{midred}88.0
& \cellcolor{darkred}99.0
& \cellcolor{darkred}100.0
& \cellcolor{darkred}97.0
& \cellcolor{lightred}62.0
& \cellcolor{lightred}56.0
& \cellcolor{lightred}60.0
& \cellcolor{midred}72.0
& \cellcolor{lightblue}53.0
& \cellcolor{lightred}60.0\\

Claude 3.7 Sonnet*
& \cellcolor{lightred}56.0
& \cellcolor{midred}72.0

& \cellcolor{lightblue}60.0
& \cellcolor{lightred}76.0

& \cellcolor{midred}82.0
& \cellcolor{lightblue}63.0

& \cellcolor{midred}70.0
& \cellcolor{lightred}60.0

& \cellcolor{midblue}35.0
& \cellcolor{midblue}38.4

& \cellcolor{midblue}22.0
& \cellcolor{lightblue}49.0
\\

Gemini-2.5-pro*
& \cellcolor{darkred}97.0
& \cellcolor{darkred}90.0

& \cellcolor{midred}84.0
& \cellcolor{lightred}78.0

& \cellcolor{darkred}100.0
& \cellcolor{darkred}99.0

& \cellcolor{lightblue}48.0
& \cellcolor{lightred}62.0

& \cellcolor{lightblue}54.0
& \cellcolor{midred}65.0

& \cellcolor{midblue}35.0
& \cellcolor{midblue}37.0
\\

\midrule
\textbf{Random} 
& 25.0 & 25.0 
& 50.0 & 50.0 
& 50.0 & 50.0 
& 25.0 & 25.0 
& 33.3 & 33.3 
& 33.3 & 33.3\\

\textbf{Human}
& 90.0 & 100.0
& 98.0 & 100.0
& 100.0 & 86.0
& 92.0 & 92.0
& 80.0 & 82.0
& 86.0 & 90.0 \\

\bottomrule
\end{tabular}
\endgroup
}
\end{minipage}
\vspace{-10pt}
\end{table*}

%% file: tables/perception_mq_results.tex
\begin{table*}[t]
\centering
\scalebox{0.7}{
\begingroup
\renewcommand{\arraystretch}{0.8}
\setlength{\tabcolsep}{2pt}
\hspace{-10pt}
\begin{tabular}{l cccc cccccccc ccccc}
\toprule
& \multicolumn{4}{c}{\textbf{Vision}}
& \multicolumn{8}{c}{\textbf{Motion}}
& \multicolumn{5}{c}{\textbf{Quantity}} \\
\cmidrule(lr){2-5}\cmidrule(lr){6-13}\cmidrule(lr){14-18}
\textbf{Model}
& \multicolumn{2}{c}{\textbf{V-C}}
& \multicolumn{1}{c}{\textbf{V-S}}
& \multicolumn{1}{c}{\textbf{V-M}}
& \multicolumn{2}{c}{\textbf{MS}}
& \multicolumn{2}{c}{\textbf{MD}}
& \multicolumn{2}{c}{\textbf{MI}}
& \multicolumn{2}{c}{\textbf{MT}}
& \multicolumn{2}{c}{\textbf{DQ}}
& \multicolumn{1}{c}{\textbf{CQ}}
& \multicolumn{2}{c}{\textbf{RQ}} \\
\cmidrule(lr){2-3} \cmidrule(lr){4-4} \cmidrule(lr){5-5}
\cmidrule(lr){6-7} \cmidrule(lr){8-9} \cmidrule(lr){10-11} \cmidrule(lr){12-13}
\cmidrule(lr){14-15} \cmidrule(lr){16-16} \cmidrule(lr){17-18}
& \textbf{Mani.} & \textbf{TDW}
& \textbf{TDW} 
& \textbf{TDW}
& \textbf{Mani.} & \textbf{TDW}
& \textbf{Mani.} & \textbf{TDW}
& \textbf{Mani.} & \textbf{TDW}
& \textbf{Mani.} & \textbf{TDW}
& \textbf{Mani.} & \textbf{TDW}
& \textbf{TDW}
& \textbf{Mani.} & \textbf{TDW} \\

\midrule
\rowcolor[HTML]{E6ECF2}
\textit{Open-source Models} &&&&&&&&&&&&&&&&& \\
NVILA 
& \cellcolor{darkred}94.6  
& \cellcolor{darkred}88.0  
& \cellcolor{darkred}98.2  
& \cellcolor{midblue}56.3  
& \cellcolor{midblue}34.2  
& \cellcolor{lightblue}45.1  
& \cellcolor{lightblue}52.6  
& \cellcolor{lightblue}53.5  
& \cellcolor{midblue}33.3  
& \cellcolor{midblue}35.3  
& \cellcolor{midblue}24.4  
& \cellcolor{midblue}14.8  
& \cellcolor{lightred}57.8  
& \cellcolor{midred}81.4  
& \cellcolor{lightred}63.9  
& \cellcolor{midred}73.5  
& \cellcolor{lightred}58.0  
\\

QWen2-VL-72b
& \cellcolor{darkred}97.2 
& \cellcolor{darkred}88.6
& \cellcolor{darkred}99.0
& \cellcolor{lightblue}63.7
& \cellcolor{midblue}33.3 
& \cellcolor{lightred}58.8
& \cellcolor{darkred}85.4 
& \cellcolor{darkred}85.4
& \cellcolor{midred}71.1 
& \cellcolor{midred}84.4
& \cellcolor{midblue}29.5 
& \cellcolor{midblue}21.5
& \cellcolor{midred}75.7 
& \cellcolor{midred}83.1
& \cellcolor{lightred}75.5
& \cellcolor{midred}75.5 
& \cellcolor{midred}74.2 \\

QWen2.5-VL-72b
& \cellcolor{darkred}94.7 
& \cellcolor{darkred}89.4
& \cellcolor{darkred}91.5
& \cellcolor{midblue}51.4
& \cellcolor{lightblue}47.6 
& \cellcolor{midred}74.7
& \cellcolor{lightred}69.8 
& \cellcolor{midred}71.9
& \cellcolor{lightred}59.7 
& \cellcolor{lightred}66.8
& \cellcolor{midblue}28.9 
& \cellcolor{midblue}27.7
& \cellcolor{midred}73.1 
& \cellcolor{midred}84.7
& \cellcolor{midred}88.0
& \cellcolor{midred}75.9 
& \cellcolor{midred}71.3 \\

InternVL2.5-78b
& \cellcolor{darkred}95.4 
& \cellcolor{darkred}89.3
& \cellcolor{darkred}95.7
& \cellcolor{midblue}53.9
& \cellcolor{midblue}23.5 
& \cellcolor{lightblue}55.3
& \cellcolor{darkred}88.5 
& \cellcolor{darkred}92.1
& \cellcolor{midred}72.2 
& \cellcolor{darkred}85.6
& \cellcolor{midblue}31.3 
& \cellcolor{midblue}25.6
& \cellcolor{midred}76.4 
& \cellcolor{midred}81.4
& \cellcolor{lightred}76.7
& \cellcolor{midred}72.1 
& \cellcolor{lightred}69.2 \\

Llama 3.2 vision-90b
& \cellcolor{darkred}95.2  
& \cellcolor{darkred}87.0  
& \cellcolor{darkred}99.6  
& \cellcolor{midblue}50.0  
& \cellcolor{midblue}27.4  
& \cellcolor{midblue}42.4  
& \cellcolor{lightblue}51.4  
& \cellcolor{midblue}36.1  
& \cellcolor{midblue}23.7  
& \cellcolor{midblue}23.8  
& \cellcolor{midblue}23.8  
& \cellcolor{midblue}25.9  
& \cellcolor{lightred}66.5  
& \cellcolor{midred}70.8  
& \cellcolor{midred}72.0  
& \cellcolor{midred}70.6  
& \cellcolor{lightred}67.4  
\\

LLaVA-OneVision
& \cellcolor{darkred}95.2 
& \cellcolor{darkred}88.8
& \cellcolor{darkred}98.4
& \cellcolor{lightred}71.2
& \cellcolor{lightblue}45.2 
& \cellcolor{midblue}40.5
& \cellcolor{lightblue}54.1 
& \cellcolor{lightblue}53.1
& \cellcolor{midblue}33.7 
& \cellcolor{midblue}25.5
& \cellcolor{midblue}26.1 
& \cellcolor{midblue}14.5
& \cellcolor{lightred}67.5 
& \cellcolor{midred}76.9
& \cellcolor{midred}81.1
& \cellcolor{midred}74.0 
& \cellcolor{midred}76.8 \\

\midrule

\rowcolor[HTML]{E6ECF2}
\textit{Closed-source Models} &&&&&&&&&&&&&&&&& \\

GPT-4o-mini
& \cellcolor{darkred}98.1  
& \cellcolor{darkred}87.8  
& \cellcolor{darkred}99.7  
& \cellcolor{midblue}50.0  
& \cellcolor{midblue}24.1  
& \cellcolor{midblue}36.7  
& \cellcolor{midblue}32.5  
& \cellcolor{midblue}37.8  
& \cellcolor{lightblue}53.9  
& \cellcolor{lightblue}51.4  
& \cellcolor{midblue}26.6  
& \cellcolor{midblue}17.1  
& \cellcolor{midblue}31.3  
& \cellcolor{lightred}60.1  
& \cellcolor{midred}73.9  
& \cellcolor{lightred}60.1  
& \cellcolor{lightblue}54.5  
\\

GPT-4o
& \cellcolor{darkred}98.6 
& \cellcolor{darkred}88.0
& \cellcolor{darkred}98.7
& \cellcolor{midblue}55.6
& \cellcolor{midblue}24.2 
& \cellcolor{lightblue}45.3
& \cellcolor{lightblue}39.6 
& \cellcolor{lightred}69.4
& \cellcolor{midblue}34.7 
& \cellcolor{midblue}38.8
& \cellcolor{midblue}29.9 
& \cellcolor{midblue}28.9
& \cellcolor{lightred}57.0 
& \cellcolor{lightred}61.3
& \cellcolor{midred}87.3
& \cellcolor{lightred}64.6 
& \cellcolor{lightred}57.8 \\

Claude 3.5 Sonnet
& \cellcolor{lightblue}42.1
& \cellcolor{lightred}62.4
& \cellcolor{darkred}96.2
& \cellcolor{midblue}50.0
& \cellcolor{midblue}35.8 
& \cellcolor{lightred}69.2
& \cellcolor{lightred}55.7 
& \cellcolor{darkred}83.8
& \cellcolor{lightblue}43.9 
& \cellcolor{lightred}52.6
& \cellcolor{midblue}24.8 
& \cellcolor{lightblue}41.7
& \cellcolor{midred}76.3 
& \cellcolor{midred}73.9
& \cellcolor{midblue}43.3
& \cellcolor{lightred}65.0 
& \cellcolor{lightblue}49.6 \\

Gemini-1.5-flash
& \cellcolor{darkred}98.3  
& \cellcolor{darkred}88.0  
& \cellcolor{darkred}98.7  
& \cellcolor{midblue}50.0  
& \cellcolor{midblue}36.2  
& \cellcolor{midblue}43.3  
& \cellcolor{lightblue}49.4  
& \cellcolor{midred}76.2  
& \cellcolor{midblue}25.2  
& \cellcolor{midblue}34.4  
& \cellcolor{midblue}24.3  
& \cellcolor{midblue}14.5  
& \cellcolor{midred}76.1  
& \cellcolor{midred}81.8  
& \cellcolor{midred}80.6  
& \cellcolor{lightred}69.2  
& \cellcolor{lightred}66.2  
\\

Gemini-1.5-pro
& \cellcolor{darkred}99.1 
& \cellcolor{darkred}86.8
& \cellcolor{darkred}89.2
& \cellcolor{midblue}50.0
& \cellcolor{midblue}36.1 
& \cellcolor{lightred}58.8
& \cellcolor{lightred}56.0 
& \cellcolor{lightred}66.1
& \cellcolor{midblue}30.0 
& \cellcolor{midblue}35.2
& \cellcolor{midblue}24.2 
& \cellcolor{midblue}33.8
& \cellcolor{darkred}84.9 
& \cellcolor{darkred}83.1
& \cellcolor{lightred}79.8
& \cellcolor{midred}73.0 
& \cellcolor{lightred}69.1 \\

GPT-4.5*
& \cellcolor{darkred}100.0
& \cellcolor{darkred}87.0
& \cellcolor{darkred}99.0
& \cellcolor{midblue}51.0
& \cellcolor{midblue}25.0
& \cellcolor{lightred}57.0
& \cellcolor{midred}75.0
& \cellcolor{darkred}92.0
& \cellcolor{midred}79.0
& \cellcolor{darkred}88.0
& \cellcolor{midblue}28.0
& \cellcolor{lightred}61.0
& \cellcolor{lightred}61.0
& \cellcolor{darkred}87.0
& \cellcolor{darkred}89.0
& \cellcolor{lightred}76.0
& \cellcolor{midred}70.0 \\

OpenAI o3*
& \cellcolor{darkred}100.0
& \cellcolor{darkred}87.0
& \cellcolor{darkred}99.0
& \cellcolor{midblue}54.0
& \cellcolor{midblue}27.0
& \cellcolor{lightred}63.0
& \cellcolor{darkred}89.0
& \cellcolor{darkred}100.0
& \cellcolor{darkred}83.0
& \cellcolor{darkred}89.0
& \cellcolor{lightred}48.0
& \cellcolor{midred}70.0
& \cellcolor{darkred}85.0
& \cellcolor{darkred}90.0
& \cellcolor{darkred}96.0
& \cellcolor{darkred}93.0
& \cellcolor{darkred}92.0\\

Claude 3.7 Sonnet*
& \cellcolor{midblue}27.3
& \cellcolor{midred}76.0

& \cellcolor{darkred}98.0
& \cellcolor{midblue}53.0

& \cellcolor{midblue}32.0
& \cellcolor{lightred}61.0
& \cellcolor{lightblue}43.8
& \cellcolor{darkred}85.7
& \cellcolor{midblue}34.0
& \cellcolor{lightblue}46.0
& \cellcolor{midblue}22.0
& \cellcolor{midblue}25.0
& \cellcolor{midred}72.0
& \cellcolor{lightred}69.0
& \cellcolor{midblue}49.0
& \cellcolor{lightred}69.0
& \cellcolor{lightblue}49.0\\

Gemini-2.5-pro*
& \cellcolor{darkred}100.0
& \cellcolor{darkred}87.0
& \cellcolor{darkred}99.0
& \cellcolor{midblue}55.0
& \cellcolor{midblue}42.0
& \cellcolor{lightred}68.0
& \cellcolor{lightblue}51.0
& \cellcolor{darkred}98.0
& \cellcolor{lightblue}46.0
& \cellcolor{lightred}60.0
& \cellcolor{midblue}36.0
& \cellcolor{lightred}51.0
& \cellcolor{midred}81.0
& \cellcolor{darkred}92.0
& \cellcolor{darkred}89.0
& \cellcolor{darkred}92.0 
& \cellcolor{darkred}93.0\\

\midrule
\textbf{Random} 
& 25.0 & 25.0
& 25.0
& 50.0
& 33.3 & 33.3 
& 25.0 & 25.0
& 25.0 & 25.0
& 25.0 & 25.0
& 25.0 & 25.0
& 50.0
& 33.3 & 33.3\\

\textbf{Human}
& 100.0 & 88.0
& 100.0
& 84.0
& 84.0 & 90.0
& 100.0 & 98.0
& 96.0 & 98.0
& 76.0 & 100.0
& 98.0 & 100.0
& 98.0
& 98.0 & 100.0 \\

\bottomrule
\end{tabular}
\endgroup}
\vspace{-5pt}
\caption{Results on the Perception tasks in our \benchmark{}, reported as accuracies (\%). Models are evaluated in two simulators \uline{Mani}Skill and \uline{T}hree\uline{DW}orld). Cell shades indicate different performance levels (\textcolor{darkred}{dark red} indicates proficient performance; \textcolor{darkblue}{dark blue} indicates performance close to random, and the lighter intermediate shades represent levels in between). \textbf{Random} and \textbf{Human} provide reference baselines. * denotes models evaluated on the subset (100 instances) of each dataset.}
\vspace{-15pt}
\label{tab:perception_main}
\end{table*}

%% file: tables/prediction_results_full.tex
\begin{table*}[ht]
\begin{minipage}{0.25\textwidth}
\centering
\includegraphics[width=1.0\textwidth]{Figures/results_legend.pdf}
\includegraphics[width=1.0\textwidth]{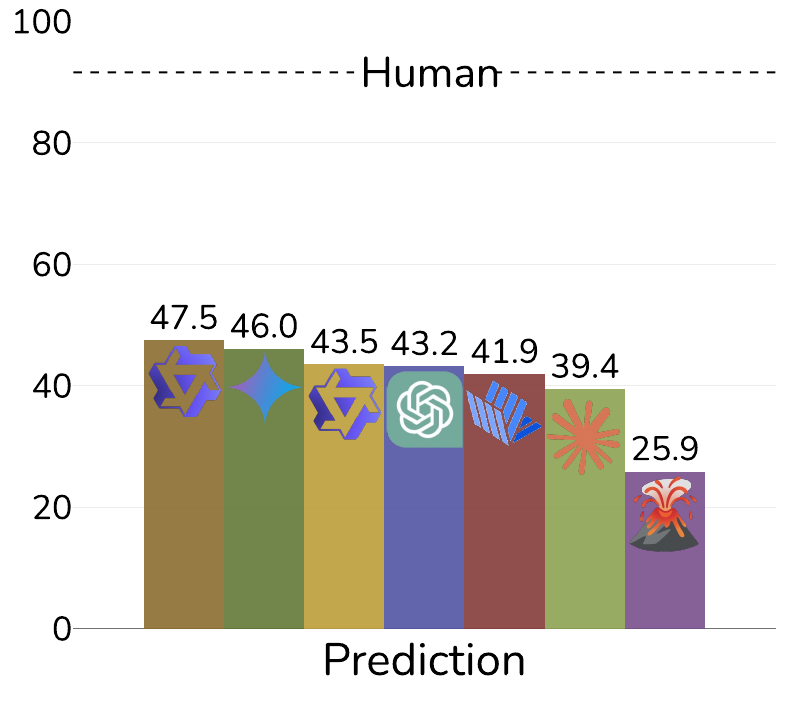}
\end{minipage}
\hspace{-10pt}
\begin{minipage}{0.6\textwidth}
\scalebox{0.625}{
\begingroup
\renewcommand{\arraystretch}{0.8}
\setlength{\tabcolsep}{1.4pt}
\begin{tabular}{lcccccccccccccccc}
\toprule
\multirow{3}{*}{\textbf{Model}} & \multicolumn{8}{c}{\textbf{Mechanistic Simulation}} 
& \multicolumn{4}{c}{\textbf{Transitivity}} 
& \multicolumn{3}{c}{\textbf{Compositionality}} \\
\cmidrule(lr){2-9}\cmidrule(lr){10-13}\cmidrule(lr){14-16} 
& \textbf{\shortstack{Car.}}  
& \textbf{\shortstack{Hab.}}
& \textbf{\shortstack{Phys.}}
& \textbf{\shortstack{Phys.}} 
& \textbf{\shortstack{Phys.}} 
& \textbf{\shortstack{Mani.}} 
& \textbf{\shortstack{Mani.}} 
& \textbf{\shortstack{Mani.}}
& \textbf{\shortstack{Car.}}  
& \textbf{\shortstack{Hab.}}  
& \textbf{\shortstack{Mani.}} 
& \textbf{\shortstack{Mani.}}
& \textbf{\shortstack{TDW}}  
& \textbf{\shortstack{Mani.}} 
& \textbf{\shortstack{Mani.}} \\
& \textbf{\texttt{nav}}  
& \textbf{\texttt{nav}}
& \textbf{\texttt{coll}}
& \textbf{\texttt{slide}}
& \textbf{\texttt{drop}} 
& \textbf{\texttt{drop}} 
& \textbf{\texttt{lift}} 
& \textbf{\texttt{push}}
& \textbf{\texttt{nav}}  
& \textbf{\texttt{nav}}  
& \textbf{\texttt{pu-pi}} 
& \textbf{\texttt{pi-ro}}
& \textbf{\texttt{coll}}  
& \textbf{\texttt{push}} 
& \textbf{\texttt{lift}} \\

\midrule
\rowcolor[HTML]{E6ECF2}
\textit{Open-source Models} & & & & & & & & & & & & & & &\\
NVILA
& \cellcolor{midblue}{31.3}
& \cellcolor{midblue}{16.7}
& \cellcolor{midblue}{28.4}
& \cellcolor{lightblue}{44.6}
& \cellcolor{lightblue}{44.7}
& \cellcolor{lightblue}{48.9}
& \cellcolor{midblue}{33.8}
& \cellcolor{midblue}{26.4}
& \cellcolor{midblue}{32.4}
& \cellcolor{midblue}{4.2}
& \cellcolor{midblue}{29.1}
& \cellcolor{midblue}{26.4}
& \cellcolor{midblue}{26.6}
& \cellcolor{midblue}{25.5}
& \cellcolor{midblue}{31.8}
\\

Qwen2-VL-72b
& \cellcolor{lightred}62.0
& \cellcolor{lightred}65.3
& \cellcolor{midblue}30.4
& \cellcolor{lightblue}40.3
& \cellcolor{lightred}59.2
& \cellcolor{darkred}95.3
& \cellcolor{darkred}91.9
& \cellcolor{midblue}20.0
& \cellcolor{lightblue}50.1
& \cellcolor{lightblue}43.8
& \cellcolor{lightblue}55.7
& \cellcolor{midblue}31.2
& \cellcolor{midblue}24.3
& \cellcolor{lightblue}51.3
& \cellcolor{lightblue}41.8 \\

Qwen2.5-VL-72b
& \cellcolor{midred}76.6
& \cellcolor{midred}70.9
& \cellcolor{midblue}36.5
& \cellcolor{lightblue}51.9
& \cellcolor{lightred}55.0
& \cellcolor{darkred}88.9
& \cellcolor{darkred}85.5
& \cellcolor{midblue}31.9
& \cellcolor{midred}61.1
& \cellcolor{midblue}34.7
& \cellcolor{midblue}24.4
& \cellcolor{midblue}30.9
& \cellcolor{midblue}21.6
& \cellcolor{midblue}34.8
& \cellcolor{midblue}35.2\\

InternVL2.5-78b
& \cellcolor{lightred}57.9
& \cellcolor{lightblue}54.6
& \cellcolor{midblue}24.9
& \cellcolor{lightblue}41.3
& \cellcolor{lightblue}51.3
& \cellcolor{midred}73.7
& \cellcolor{lightred}60.3
& \cellcolor{midblue}28.1
& \cellcolor{lightblue}51.1
& \cellcolor{midblue}31.9
& \cellcolor{midblue}35.6
& \cellcolor{midblue}34.7
& \cellcolor{lightblue}40.1
& \cellcolor{lightblue}47.0
& \cellcolor{midblue}27.8 \\

Llama-3.2-90b
& \cellcolor{midblue}{24.7}
& \cellcolor{midblue}{33.3}
& \cellcolor{midblue}{24.0}
& \cellcolor{midblue}{25.4}
& \cellcolor{midblue}{24.7}
& \cellcolor{midblue}{23.7}
& \cellcolor{midblue}{26.3}
& \cellcolor{midblue}{23.1}
& \cellcolor{midblue}{23.9}
& \cellcolor{midblue}{33.6}
& \cellcolor{midblue}{25.6}
& \cellcolor{midblue}{20.8}
& \cellcolor{midblue}{25.9}
& \cellcolor{midblue}{24.1}
& \cellcolor{midblue}{25.3}
\\

LLaVA-OneVision
& \cellcolor{midblue}19.0
& \cellcolor{lightblue}57.9
& \cellcolor{midblue}25.8
& \cellcolor{midblue}20.5
& \cellcolor{midblue}19.4
& \cellcolor{midblue}26.1
& \cellcolor{midblue}30.8
& \cellcolor{midblue}25.1
& \cellcolor{midblue}22.6
& \cellcolor{midblue}26.7
& \cellcolor{midblue}25.4
& \cellcolor{midblue}22.4
& \cellcolor{midblue}22.4
& \cellcolor{midblue}25.0
& \cellcolor{midblue}28.2 \\

\midrule
\rowcolor[HTML]{E6ECF2}
\textit{Closed-source Models} & & & & & & & & & & & & & & &\\

GPT-4o
& \cellcolor{lightred}61.6
& \cellcolor{lightred}69.6
& \cellcolor{midblue}39.2
& \cellcolor{midblue}37.6
& \cellcolor{lightblue}49.4
& \cellcolor{midred}83.9
& \cellcolor{midred}71.2
& \cellcolor{midblue}30.3
& \cellcolor{lightblue}42.6
& \cellcolor{lightblue}43.2
& \cellcolor{lightblue}42.4
& \cellcolor{lightblue}45.5
& \cellcolor{midblue}22.8
& \cellcolor{midblue}33.8
& \cellcolor{midblue}35.6 \\

GPT-4o-mini
& \cellcolor{lightblue}{45.8}
& \cellcolor{midblue}{34.0}   
& \cellcolor{midblue}{25.8}   
& \cellcolor{midblue}{32.9}    
& \cellcolor{lightblue}{49.9} 
& \cellcolor{midred}{77.7}     
& \cellcolor{lightred}{68.2}  
& \cellcolor{midblue}{30.9}    
& \cellcolor{lightblue}{41.6} 
& \cellcolor{midblue}{24.4}    
& \cellcolor{midblue}{32.9}    
& \cellcolor{midblue}{39.4}    
& \cellcolor{midblue}{37.2}   
& \cellcolor{lightblue}{40.8}  
& \cellcolor{midblue}{30.8}    
\\

Claude 3.5 Sonnet
& \cellcolor{midblue}36.3
& \cellcolor{midblue}39.6
& \cellcolor{midblue}15.0
& \cellcolor{midblue}22.1
& \cellcolor{midblue}36.7
& \cellcolor{darkred}86.3
& \cellcolor{lightred}57.7
& \cellcolor{midblue}28.0
& \cellcolor{midblue}35.3
& \cellcolor{midblue}27.5
& \cellcolor{midblue}36.3
& \cellcolor{midblue}37.5
& \cellcolor{midblue}39.0
& \cellcolor{lightblue}41.9
& \cellcolor{lightblue}51.0 \\

Gemini-1.5-flash
& \cellcolor{midblue}{36.8}  
& \cellcolor{lightblue}{41.1}
& \cellcolor{midblue}{27.3}   
& \cellcolor{midblue}{30.2}  
& \cellcolor{lightblue}{42.2} 
& \cellcolor{midred}{75.7}  
& \cellcolor{lightred}{65.8} 
& \cellcolor{midblue}{18.4}   
& \cellcolor{lightblue}{40.7}
& \cellcolor{midblue}{39.0}  
& \cellcolor{lightblue}{42.7} 
& \cellcolor{midblue}{37.0}   
& \cellcolor{lightblue}{40.2} 
& \cellcolor{midblue}{30.8}   
& \cellcolor{lightblue}{46.2}\\ 

Gemini-1.5-pro
& \cellcolor{lightblue}44.3
& \cellcolor{lightred}57.6
& \cellcolor{lightblue}47.5
& \cellcolor{midblue}37.6
& \cellcolor{lightred}60.1
& \cellcolor{midred}79.4
& \cellcolor{midred}76.8
& \cellcolor{midblue}33.5
& \cellcolor{lightblue}48.1
& \cellcolor{lightblue}43.3
& \cellcolor{lightblue}46.2
& \cellcolor{midblue}35.9
& \cellcolor{midblue}34.1
& \cellcolor{midblue}36.8
& \cellcolor{lightblue}49.6 \\

GPT-4.5*
& \cellcolor{midred}71.0
& \cellcolor{midblue}34.0
& \cellcolor{lightred}55.0
& \cellcolor{midred}72.0
& \cellcolor{lightred}56.0
& \cellcolor{darkred}86.0
& \cellcolor{lightred}59.0
& \cellcolor{lightblue}40.0
& \cellcolor{lightblue}47.0
& \cellcolor{midblue}34.0
& \cellcolor{lightblue}44.0
& \cellcolor{midblue}35.0
& \cellcolor{midblue}32.0
& \cellcolor{lightblue}41.0
& \cellcolor{midblue}39.0\\

OpenAI o3*
& \cellcolor{darkred}85.0
& \cellcolor{darkred}87.0
& \cellcolor{lightred}60.0
& \cellcolor{midred}78.0
& \cellcolor{darkred}79.0
& \cellcolor{midred}80.0
& \cellcolor{midred}83.0
& \cellcolor{lightblue}50.0
& \cellcolor{midred}61.0
& \cellcolor{lightred}51.0
& \cellcolor{lightblue}42.0
& \cellcolor{lightblue}48.0
& \cellcolor{midblue}33.0
& \cellcolor{midblue}37.0
& \cellcolor{midblue}37.0\\

Claude 3.7 Sonnet*
& \cellcolor{lightred}{59.0}
& \cellcolor{lightred}{51.0}
& \cellcolor{midblue}{38.5}
& \cellcolor{lightblue}{49.0}
& \cellcolor{midred}{67.7}
& \cellcolor{darkred}{96.0}
& \cellcolor{lightred}{56.4}
& \cellcolor{midblue}{39.4}

& \cellcolor{lightblue}{44.0}
& \cellcolor{midblue}{26.0}
& \cellcolor{midblue}{34.3}
& \cellcolor{lightblue}{48.4}

& \cellcolor{lightblue}{41.5}
& \cellcolor{midblue}{29.0}
& \cellcolor{lightblue}{49.5}
\\

Gemini-2.5-pro*
& \cellcolor{darkred}{81.0}
& \cellcolor{midred}{77.0}
& \cellcolor{lightred}{53.0}
& \cellcolor{lightred}{68.0}
& \cellcolor{midred}{73.0}

& \cellcolor{darkred}{87.0}
& \cellcolor{midred}{76.0}
& \cellcolor{lightblue}{44.0}

& \cellcolor{lightred}{66.0}
& \cellcolor{lightred}{67.0}
& \cellcolor{lightblue}{42.0}
& \cellcolor{lightred}{54.0}

& \cellcolor{midblue}{38.0}
& \cellcolor{lightblue}{47.0}
& \cellcolor{lightred}{57.0}
\\

\midrule
\textbf{Random} 
& 25.0 & 25.0 & 25.0 & 25.0 & 25.0 & 25.0 & 25.0 & 25.0
& 25.0 & 25.0 & 25.0 & 25.0
& 25.0 & 25.0 & 25.0\\

\textbf{Human}
& 98.0 & 98.0 & 100.0 & 98.0 & 86.0 & 100.0 & 100.0 & 100.0  
& 78.0 & 90.0 & 80.0 & 82.0
& 84.0 & 88.0 & 100.0  \\

\bottomrule
\end{tabular}
\endgroup}
\end{minipage}
\vspace{-5pt}
\caption{Results of the Prediction tasks in WM Benchmark (\%). Our mechanistic simulation tasks cover three categories: intuitive physics (e.g., \uline{drop}, \uline{slide} \uline{coll}ision), agent manipulation (e.g., \uline{lift}, \uline{push}), and \uline{nav}igation (e.g., turn left, move forward). * denotes models evaluated on the subset (100 instances) of each dataset. }
\vspace{-15pt}
\label{tab:prediction_main}
\end{table*}

%% file: MainSections/04-Experiments.tex
\vspace{-5pt}
\section{Experiments}
\label{sec:experiment}
\vspace{-2pt}

\subsection{Experimental Setup}

\paragraph{Evaluated Models.}
We evaluate a range of state-of-the-art VLMs on \benchmark, including {\bf closed-source models}: Gemini-1.5-pro, Gemini-1.5-flash \citep{gemini2024}, GPT-4o (\texttt{gpt-4o-2024-08-06}), GPT-4o-mini \citep{openai2024gpt4}, Claude 3.5 Sonnet \citep{anthropic2024claude}, and {\bf open-source models}: Qwen2-VL (72B) \citep{Qwen-VL}, Qwen2.5-VL (72B) \citep{qwen2.5}, InternVL2.5 (78B) \citep{internvl2.5}, LLaVA-OneVision (72B) \citep{llava_onevision}, NVILA (15B) \citep{nvila}, and Llama 3.2-Vision (90B) \citep{metaAI2024blog}.

We used the same system prompt (see Appendix \ref{appendix:prompting_framework}) and greedy decoding for all questions to maintain consistent output formatting. 
We evaluate model performance by comparing the parsed labels from model outputs to the ground-truth labels, and skip the instances where model outputs failed to be parsed by our template. 

\vspace*{-5pt}
\paragraph{Human Evaluation.}
For each subtask, we randomly sample 50 questions and employ Amazon Mechanic Turk for human evaluations. 
We asked 3 evaluators for each question and finalized the labels via majority voting. 
The inter-rater agreement is measured by Fleiss' kappa on 50 samples for each task. 
All tasks are above moderate agreement (Fleiss' $k > 0.4$).
Detailed instructions are available in Appendix \ref{appendix:human_eval}).

\vspace*{-3pt}
\subsection{Main Results on Perception Tasks}
Table~\ref{tab:perception_main} presents the results for perception tasks. 
In fact, closed-source VLMs do not exhibit an overwhelming advantage over open-source models in terms of perceptual capabilities.
Across the 5 dimensions, Qwen2-VL achieves the highest overall performance with an average accuracy of 67.7\%. 
All models still fall behind human-level perception, which is near-perfect or substantially more accurate across all the perception tasks.

\vspace*{-5pt}
\paragraph{Recommendation 1: Improve 3D Perception.}
Spatial tasks require grounding and reasoning over 3D semantics, including constructing robust scene representations from limited views. 
Even the most advanced models achieve less than 60\% accuracy in spatial positioning tasks.
These results suggest that current VLMs struggle to form robust internal 3D representations, in line with previous findings~\citep{el2024probing,zhang2025do}. 
We recommend future VLMs transition from relying solely on 2D semantics to incorporating 3D priors or explicit 3D representations.

\vspace*{-5pt}
\paragraph{Recommendation 2: Improve Temporal and Motion Understanding.}
We found that models struggle with coherent temporal representations across consecutive frames, as reflected in their low performance on temporal extension (TE). 
In contrast, they perform much better on tasks relying on a subset of frames, such as temporal positioning (TP). 
Similarly, while models perform relatively well on motion detection (MD), they exhibit \textit{near-random performance} on motion trajectory (MT), which demands a thorough understanding of consecutive states. 
These results suggest that current VLMs still struggle to perceive and form dynamic scene representations.
We recommend future VLMs incorporate temporal and motion priors by leveraging the rich visual dynamics in videos~\citep{song2024moviechat,ko2024learning}.

\vspace*{-3pt}
\subsection{Main Results on Prediction Tasks}

Table~\ref{tab:prediction_main} presents the results for prediction tasks, which generally pose greater challenges for VLMs than perception tasks. 
Qwen2-VL achieves the highest average accuracy of 47.5\%. 
Similar to perception tasks, open-source VLMs perform on par with closed-source ones, and all VLMs fall behind human performance by a large margin.

\vspace*{-5pt}
\paragraph{Recommendation 3: Improve Cause-and-Effect Understanding.}
Our prediction tasks cover intuitive physics (e.g., drop, collision) and agent-initiated actions like navigation (e.g., turn left, move forward), and manipulation (e.g., lift, push). 
We find that VLMs struggle with predicting the post-condition of physical transitions and manipulation actions. 
For instance, on the ManiSkill simulator, Qwen2-VL achieves 95.3\% accuracy in predicting the outcomes of dropping and 91.4\% in lifting. 
However, its performance on pushing objects in the same environment is close to random, and its accuracy in predicting dropping outcomes on Physion decreases to 59.2\%.
These results suggest that current VLMs still struggle to reliably predict the cause and effect of physical processes and actions~\citep{gao2018action}.

\vspace*{-5pt}
\paragraph{Recommendation 4: Improve Transitive and Compositional World Modeling.}
Our results also reveal that VLMs do poorly in transitive and compositional inference when reasoning over sequential or concurrent actions. 
In transitive inference tasks, models consistently underperform, with even the best achieving only 43.8\% on the multi-step navigation task in Habitat, which is far below the human accuracy of 90.0\%. 
Similarly, in compositional inference tasks, the gap remains substantial: the best models reach only 40.2\% on the collision prediction task in TDW and 51.3\% on the manipulation task in ManiSkill, compared to the human performance of 84.0\% and 88.0\%.

\subsection{Performance of Frontier VLMs}
We extend our evaluation to include the most recent frontier VLMs \cite{Kavukcuoglu2025Gemini, OpenAI2025o3o4minisystemcard, OpenAI2025GPT45SystemCard, Anthropic2025ClaudeSonnetSystemCard}, including visual reasoning models such as OpenAI o3. Due to computational constraints and API limitations, we evaluate these frontier models on a subset of 100 instances per task.
\paragraph{Frontier Models Reach Human Parity in Static Perception.}
Frontier VLMs (e.g., Gemini-2.5-Pro, o3, GPT-4.5) demonstrate clear improvements over previous state-of-the-art models across static perception tasks. On the Size Comparison (SE-O) task in ThreeDWorld, frontier models achieve 92-100\% accuracy, up from GPT-4o's 88\%, even exceeding the 93\% human baseline. Similarly, Relative Position (SR) performance improves from 71-73\% (LlaVA-OneVision, QWen2-VL) to 90-97\% (Gemini-2.5-Pro). 
\paragraph{Notable Gains in Mechanistic Simulation but Gaps Remain.}
We also observe substantial improvements from these models in mechanistic simulation tasks. In Carla Navigation task, accuracy rises from 60\% (GPT-4o) to 81-85\% for Gemini-2.5-Pro and OpenAI o3, though still below 98\% human performance. Intuitive physics tasks (Physion Collision, Slide) advance from 47-60\% to 60-79\%. 
\paragraph{Persistent Challenges in Spatial, Temporal, and Compositional Inference.}
Depite these performance gains, certain spatial and temporal perception tasks remain challenging. Multi-view spatial reasoning (SP) for these frontier models does not exceed 70\%, well under the 90\%+ human level. Temporal Extension (TE) shows minimal improvement, with only 7.3\% gains from GPT-4.5. Advanced prediction tasks also shows little progress, with the highest ManiSkill Lift performance (57\%, Gemini-2.5-Pro) significantly lagging human capabilities.

%% file: MainSections/05-Discussion.tex
\section{Further Analyses and Discussions}
\label{sec:discussion}

\begin{figure*}
    \centering
    \includegraphics[width=\linewidth]{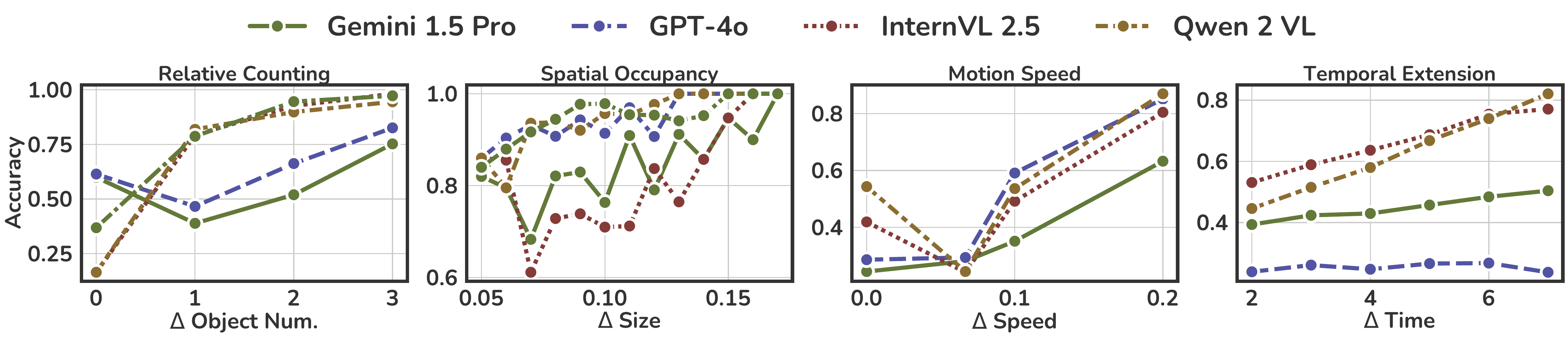}
    \vspace{-20pt}
    \caption{Model performance with respect to increasing stimulus differences, as discussed in Section~\ref{sec:vlm-stimulus}. Here, $\Delta_{\text{Object Num.}}$, $\Delta_{\text{Size}}$, $\Delta_{\text{Speed}}$, and $\Delta_{\text{Time}}$ represent the differences in the number of objects, object sizes, object speeds, and object movement durations, respectively.}
    \vspace{-6pt}
    \label{fig:differentiability}
\end{figure*}

\subsection{VLMs fail to represent different world attributes robustly and independently}

World models should learn disentangled representations of key perceptual dimensions, like color or spatial position, for flexible compositional reasoning and categorical distinctions.\footnote{Human perception also relies on interdependent encoding of visual properties, e.g., object attributes are represented as statistical summaries rather than in isolation \citep{whitney2018ensemble}. However, unlike VLMs, human cognition can differentiate these summaries into discrete symbolic states, as evident in our evaluations, enabling precise world representation despite potential interdependencies.}
To assess entanglement, we perturb one dimension while keeping others constant and define the standardized Relative Entanglement (s-RE) as the normalized performance deviation between different perturbations:
\(\mathrm{s\text{-}RE} = \frac{1}{N} \sum_{i=1}^N \left|\frac{\bar{p} - p_i}{\bar{p}}\right|\), where $p_{i}$ is model performance under the $i_{th}$ perturbation and $\bar{p}$ is the average performance over all perturbations. Essentially, s-RE calculates the average relative deviation from the mean performance, reflecting model sensitivity to changes in that particular dimension.
We control 5 perceptual dimensions (\textit{color}, \textit{shape}, \textit{size}, \textit{position}, \textit{material}) and present the entanglement heatmap in Figure \ref{fig:heatmap}.
Our results show that VLMs' representations of orthogonal world attributes have significant entanglement.
Color and Shape are major sources of entanglement in multiple tasks.
For example, color entangles with the Discrete Quantity (DQ) task, where s-RE ranges from 5\% (Gemini-1.5 Pro) to 17\% (Qwen-2.5 VL). 
Other dimensions like size and absolute position also exhibit relatively modest entanglement effects. 
\input{tables/claim4_table}

\vspace{-5pt}
\subsection{VLMs are sensitive to stimulus differences, but not fine details}
\label{sec:vlm-stimulus}
As shown in Figure~\ref{fig:differentiability}, we find that VLM performance is positively correlated with the physical differentiability of stimuli, such as differences in size, speed, or moving duration. 
This aligns with the previously reported ``myopia'' in model perception~\citep{rahmanzadehgervi2024visionlanguagemodelsblind}.
Our observation suggests that VLMs can, to some extent, effectively ground language in corresponding physical attributes.
This is evidenced by their strong performance in scenarios with large stimulus differences, which would otherwise be inexplicable.
On the other hand, VLMs' perception capabilities are strongly influenced by the magnitude of the stimulus, irrespective of specific physical attributes. 
This suggests that while they perform well when distinctions are pronounced, they struggle with fine-grained, high-resolution perception, highlighting a significant gap in modeling subtle physical variations.

\vspace{-5pt}
\subsection{VLMs struggle with intuitive physics even under accurate state perception}
Accurate next-state prediction relies on two key factors: correctly representing the current state and possessing sufficient mechanistic knowledge for transition simulation. 
Our analysis reveals that models often make mistakes in perception, which may contribute to their underperformance in prediction tasks, stemming from limited world modeling capabilities or accumulated errors resulting from perception failures.
To disentangle perception from prediction, we present a further analysis of the mechanistic simulation from intuitive physics (\textit{collide}, \textit{slide}, \textit{drop}).
We retrieve instances where all models correctly answer all relevant perception questions, ensuring an accurate state representation (details provided in Appendix~\ref{appendix:perceptual queris}). 
As Table~\ref{tab:claim4} shows, performance on predicting the next state of \textit{slide} and \textit{drop} increases only marginally or even decreases for \textit{collide}, indicating that limited perception capability is not the only cause. 
Rather, models lack the foundational physical knowledge to simulate object interactions accurately. 

\vspace{-5pt}
\subsection{\benchmark from Real-World Data}
\input{tables/real}
Transferring WM-ABench's controlled settings to real-world scenarios is challenging due to the need for precise attribute manipulations (e.g., altering object colors) and counterfactual state generation (e.g., applying forces to the target object). To draw some insights on whether our simulation findings might generalize to real-world data, we repurpose existing datasets~\cite{bridgeV2,liu2024mmbench,RGB-DScenes,ActiveVisionDataset,yu2023mvimgnet,shangguan2025tomato,MotIFDataset} and recast them into our evaluation format. For each dimension, we curated approximately 50 data points and evaluated several top-performing models. 

Results in Table~\ref{tab:real} demonstrate consistent alignment between real-world and simulated performance patterns. For instance, models exhibit poor performance on Spatial Positioning (SP) and Motion Trajectory (MT), mirroring simulation trends. Conversely, they achieve near-perfect accuracy on Color and Shape recognition tasks. Additionally, Temporal Positioning performance (70–90\%) consistently exceeds Temporal Extension (50–76\%), confirming our earlier findings. These trends suggest that bias from our use of simulation data is likely to be minimal, and our findings tend to be generalizable.

%% file: tables/claim4_table.tex
\begin{table}[t!]
\centering
\scalebox{0.8}{
\begingroup
\renewcommand{\arraystretch}{0.8}
\setlength{\tabcolsep}{3pt}
\hspace{-10pt}
\begin{tabular}{lcccc}
\toprule
 & \textbf{GPT-4o} & \textbf{Gemini-1.5} & \textbf{Qwen2-VL} & \textbf{InternVL 2.5}\\
\cmidrule(lr){1-1}\cmidrule(lr){2-5}
$\Delta_{\texttt{collide}}$ & -1.09 & -0.52 & -0.62 & 0.36 \\

$\Delta_{\texttt{slide}}$ & 1.67 & 4.02 & 1.36 & 1.15 \\

$\Delta_{\texttt{drop}}$ & 3.55 & 1.50 & 1.94 & 2.95\\
\bottomrule
\end{tabular}
\endgroup}
\vspace*{-8pt}
\caption{Accuracy differences ($\Delta(\cdot)\%$) between filtered (correct state perception) and unfiltered inputs across physical transition tasks.}
\vspace*{-20pt}
\label{tab:claim4}
\end{table}

%% file: tables/real.tex
\begin{table}[t]
\hspace*{-0.7em}
\scalebox{0.65}{
\begingroup
\renewcommand{\arraystretch}{0.8}
\setlength{\tabcolsep}{2pt} 
\begin{tabular}{lcccccc@{\hspace{8pt}}cc}
\toprule
 & \multicolumn{1}{c}{\textbf{Space}}
 & \multicolumn{2}{c}{\textbf{Time}}
 & \multicolumn{2}{c}{\textbf{Vision}}
 & \multicolumn{1}{c}{\textbf{Motion}}
 & \multicolumn{2}{c}{\makecell{\textbf{Mechanistic}\\\textbf{Simulation}}}\\
\cmidrule(lr){2-2}
\cmidrule(lr){3-4}\cmidrule(lr){5-6}\cmidrule(lr){7-7}\cmidrule(lr){8-9}
\textbf{Model}
 & \textbf{SP} & \textbf{TP} & \textbf{TE}
 & \textbf{V-C} & \textbf{V-S}
 & \textbf{MT}
 & \textbf{Lift} & \textbf{Drop}\\
\midrule

\rowcolor[HTML]{E6ECF2}\multicolumn{9}{l}{\textit{Open-source Models}}\\

QWen2-VL-72b
& \cellcolor{white}32.1
& \cellcolor{white}76.7
& \cellcolor{white}63.3 
& \cellcolor{white}100.0 
& \cellcolor{white}100.0  
& \cellcolor{white}30.0 
& \cellcolor{white}53.3 
& \cellcolor{white}80.0 
\\

QWen2.5-VL-72b
& \cellcolor{white}35.3 
& \cellcolor{white}76.7
& \cellcolor{white}76.7 
& \cellcolor{white}100.0  
& \cellcolor{white}100.0  
& \cellcolor{white}23.3 
& \cellcolor{white}76.7 
& \cellcolor{white}73.3   \\

InternVL2.5-78b
& \cellcolor{white}30.0
& \cellcolor{white}90.0
& \cellcolor{white}63.3 
& \cellcolor{white}100.0 
& \cellcolor{white}100.0  
& \cellcolor{white}40.0 
& \cellcolor{white}30.0
& \cellcolor{white}50.0   \\

\midrule

\rowcolor[HTML]{E6ECF2}\multicolumn{9}{l}{\textit{Closed-source Models}}\\

Claude 3.5 Sonnet
& \cellcolor{white}26.7 
& \cellcolor{white}70.0
& \cellcolor{white}60.0 
& \cellcolor{white}100.0  
& \cellcolor{white}100.0  
& \cellcolor{white}26.7 
& \cellcolor{white}40.0 
& \cellcolor{white}66.7  
\\

Gemini-1.5-flash 
& \cellcolor{white}43.3 
& \cellcolor{white}83.3
& \cellcolor{white}60.0 
& \cellcolor{white}100.0  
& \cellcolor{white}100.0  
& \cellcolor{white}30.0 
& \cellcolor{white}36.7 
& \cellcolor{white}50.0  
\\

Gemini-1.5-pro
& \cellcolor{white}53.3 
& \cellcolor{white}80.0
& \cellcolor{white}63.3 
& \cellcolor{white}100.0  
& \cellcolor{white}97.7  
& \cellcolor{white}40.0 
& \cellcolor{white}53.3 
& \cellcolor{white}76.7  
\\

GPT-4o
& \cellcolor{white}50.0 
& \cellcolor{white}90.0
& \cellcolor{white}60.0 
& \cellcolor{white}100.0  
& \cellcolor{white}97.7  
& \cellcolor{white}26.7 
& \cellcolor{white}60.0 
& \cellcolor{white}80.0  
\\

GPT-4o-mini
& \cellcolor{white}33.3 
& \cellcolor{white}83.3
& \cellcolor{white}50.0 
& \cellcolor{white}100.0  
& \cellcolor{white}100.0  
& \cellcolor{white}30.0 
& \cellcolor{white}60.0 
& \cellcolor{white}53.3  
\\

\midrule
\textbf{Random} 
& 25.0 & 33.0
& 33.0
& 25.0
&25.0 & 25.0
& 25.0 & 25.0 \\

\bottomrule
\end{tabular}
\endgroup}
\vspace{-5pt}
\caption{Results across 8 dimensions adapted from real-world datasets into \benchmark design. We recast from real datasets and curate approximately 50 data points for each selected dimension.}
\vspace{-15pt}
\label{tab:real}
\end{table}

%% file: MainSections/06-Related_work.tex
\vspace{-5pt}
\section{Related Work}

\vspace{-5pt}
\paragraph{World Models.}

World models (WMs) predict how the current state transitions to the next, based on prior states and actions \citep{tolman1948cognitive, battaglia2013simulation}. 
Traditionally, people train frame-level video-generative models specializing in some narrow domains. 
For instance, in robotics, WMs enable model-based reinforcement learning and trajectory prediction \citep{yang2023learning,zhou2024robodreamer}; in autonomous driving \citep{wang2023driving,hu2023gaia}, they facilitate path planning; and in gaming \citep{hafner2019dream,bruce2024genie, recurrent-world-model,world-model}, they power interactive simulations. 
Meanwhile, recent work \citep{brooks2024video,kang2024far} has explored whether video-generation models can serve as world simulators that go beyond mere pixel-level synthesis. 
We instead investigate whether Vision-Language Models can capture world dynamics from large-scale training data, enabling them to function as generalist world models.

\vspace{-5pt}
\paragraph{Benchmarks for Vision Language Models.}

Previous VLM benchmarks typically take a reductionist approach, measuring a wide range of perceptual capabilities while giving limited attention to how these perceptual dimensions interact and influence one another. 
For instance, there are works focusing on \emph{visual semantics perception}, e.g., object categories, attributes, actions, agent-object interactions, emotions~\citep{liu2024mmbench,li2024mvbench,liu2023visual}, whereas others emphasize \emph{low-level visual perception}, e.g., basic attributes, line segments, optical flow~\citep{johnson2016clevr,fu2024blink,shitsukan}, \emph{spatiotemporal and motion}, e.g., geometry, event ordering, trajectories~\citep{goyal2020rel3dminimallycontrastivebenchmark,mirzaee-etal-2021-spartqa,shangguan2024tomatoassessingvisualtemporal}, or \emph{next-state prediction} (often limited to intuitive physics) \citep{bear2022physion,yi2020clevrercollisioneventsvideo}.
Compared to these efforts, our framework decomposes world modeling into atomic dimensions, offering a precise diagnosis of models' perception and prediction capabilities while establishing a clear checklist for VLMs as world models.
Due to the page limit, we leave the comprehensive comparison between \benchmark and existing benchmarks in Table~\ref{tab:benchmarks} and Appendix~\ref{sec:related_bench}.

%% file: MainSections/07-Conclusion.tex
\vspace{-5pt}
\section{Conclusion}
\vspace{-2pt}

Our study provides the first atomic evaluation of VLMs' internal world modeling abilities with a cognitively inspired framework. While VLMs excel in scenarios with pronounced differences, they struggle with 3D and dynamic perception, fail to differentiate subtle physical distinctions, and exhibit failures in understanding world transitions of transitive and compositional scenarios.

%% file: MainSections/08-Limitation.tex
\section*{Limitations}
\label{limitation}
While simulators enable compute-scalable and cheap generation of dataset questions and answers, most simulators are difficult to tune or are incapable of photo-realistic image generation. As a result, we may be evaluating VLMs on somewhat out-of-distribution data as the images do not look realistic and the majority of the image data used to train VLMs likely come from real-world videos/images. While ray tracing is used in some of the ManiSkill-generated problems, better photo-realism can be achieved if higher quality assets are used and lighting is tuned better.

\section*{Ethics Statement}
\label{ethics}

\vspace{-5pt}
\paragraph{Human Study.}
We only involve human subjects to obtain the human performance on our benchmark. 
The institution’s Institutional Review Board (IRB) considered this project exempt from ongoing review. 
All human evaluations in our study are conducted in Amazon Mechanical Turk (MTurk), with participants fairly compensated according to ethical guidelines. Further details on the evaluation process are provided in the Appendix \ref{appendix:human_eval}).

\vspace{-5pt}
\paragraph{Societal Impact.}
Our benchmark aims to provide an atomic evaluation of VLMs' fundamental WM abilities.
Since our experiments are conducted in simulated environments rather than using real-world data, our work does not raise ethical concerns related to privacy, bias, or societal impact.

\vspace{-5pt}
\paragraph{Licenses.}
We strictly adhere to the protocols governing the academic use of all VLMs. Our benchmark is designed to evaluate the performance of VLMs across various dimensions, focusing primarily on inference rather than the training of large-scale models. 
As a result, it is unlikely to have a significant environmental impact. 

\vspace{-5pt}
\paragraph{AI Assistants In Research.}
Furthermore, regarding the use of AI tools, we only used ChatGPT for grammar checking and made minor refinements to our visualization code with its assistance, which is entirely unrelated to the scientific content or findings of our work.

\section*{Acknowledgments}
We would like to thank Angela Shen, Kai Kim, Reventh Sharma, and Prathish Murugan for their contribution to early data collection, and Yi Gu, Yuheng Zha for assistance with setting up simulation environments.

%% file: Appendices/a1_framework.tex
 \section{Conceptual Framework Explained}
\label{conceptualframework}

\subsection{The Dual-Stage Model}

In general, world models (WMs) predict the future states of the world (based on observations of the past and current states) and the next action to be taken.\footnote{We regard the corner case of ``inaction'' as an element of the action space}
Formally, a WM $\theta$ recursively models the transition distribution: $ P_\theta(S_{t} \mid S_1, a_1, \ldots, S_{t-1}, a_{t-1})$.

With theoretical consistency with previous research \cite{knill2004bayesian, ha2018world, Smith2024-jb}, we decouple the world modeling process into two stages.
In the first stage, an agent encodes environmental stimuli from sensory signals into internal representations of the external world.
In the second stage, the agent performs an extrapolation into possible future states.
As time progresses and future states come into embodiment, the agent acquires ground truth data and updates its WM based on the divergence between prediction and reality.
Unlike model-free RL, where the agent optimizes its action policy to maximize utility, world modeling is primarily a supervised learning problem, optimized through error reduction.

Given this dual-stage framework, world modeling can fail for two different reasons: 
\begin{itemize}[leftmargin=10pt, topsep=1pt, noitemsep]
\item Perceptual limitations that lead to problematic representations of external states. For example, an agent may encode different colors identically or conflate object size with speed.
\item Prediction limitations that arise from insufficient mechanistic knowledge and lead to inaccurate simulations. For example, an agent may fail to follow momentum conservation or struggle with complex multi-object collisions.
\end{itemize}
Our framework addresses these challenges by considering both perception competency and prediction competency. 
We discuss how we further decompose the two stages to make this benchmark empirically possible.

\subsection{Perception Process}

Accurate prediction of future states relies on constructing precise representations of the current environment, a process known as perception \cite{goldstein1989sensation}. 
Perceptions have multitudinous dimensions: force, motion, temperature, sound, magnetics, space, time, quantity, and other agents, to name a few \cite{merleau2004world, coren2004sensation}.
Agents detect physical signals from external stimuli through their physiological sensors, converting multimodal raw inputs into physio-electronic signals. 
These signals then undergo complex bottom-up processing and functional transformations until they become recognizable to higher cognitive faculties involved in semantics and reasoning. 
Subsequently, top-down processing refines and hypothesizes representations of the world state. This bidirectional interaction continues iteratively until convergence.
Such top-down processing enables more reliable and robust representations of the world. \cite{rao1999predictive, frith1997brain, delorme2004interaction, mechelli2004bottom}
For example, animals such as crows, dolphins, and chimpanzees exhibit an understanding of object permanence, allowing them to maintain stable representations of objects even after they become obscured or disappear \cite{Baillargeon1985-uu, hoffmann2011ontogeny, Wood1980-gc, Mitroff2004-lx}.
Due to the sensory capabilities of current VLMs, we keep spatial, temporal, quantitative, visual, and motion perceptions in our framework only to accommodate their status quo.

\subsection{Prediction Process}

To formalize how next-state prediction is processed, we draw on the compositional generalization framework from \citet{dziri2023faith} and consider next-state predictions as a topologically sorted computational graph.
Given a world environment, each state can be hierarchically decomposed into a structured sequence of nodes, where a node intuitively represents an object as a vector of attributes (e.g., speed, mass, direction, color). 
The underlying rationale is as follows:
\begin{enumerate}[leftmargin=10pt, topsep=1pt, noitemsep]
    \item Any valid program can be expressed as an equivalent topologically sorted computational graph based on dependency structures;
    \item All world simulators function as programs;
    \item Therefore, all world simulators have equivalent graphs. The topological order of the graph is determined by time and world dynamics. 
\end{enumerate}
Within this framework, future state prediction involves computing values for future nodes based on historical nodes, using the current time step as a cutoff. 
We then identify three necessary and collectively sufficient conditions.

\paragraph{Atomic Mechanistic Simulation.} 
Mechanism simulation refers to the process of predicting a system's future states via representing and modeling the dynamic interactions among its component parts \cite{10.1093/acprof:oso/9780199933662.001.0001, hegarty2004mechanical, barsalou2008grounded}.
This is to be contrasted with statistical predictions, which typically rely on correlational shortcuts and bypass explicit simulations of component behaviors over time \cite{kendall1999populations}.
Mechanistic knowledge about natural laws and object-specific properties, functions, behavioral patterns, and interactive dynamics are the fuel of mechanistic simulation \cite{sep-science-mechanisms}, and are mostly learned in a posterior manner \cite{kant1934critique}.
Here we especially emphasize the atomicity of mechanistic simulation, i.e., single object motion or minimally viable object interactions. 
More complicated world dynamics are left to the compositional inference part.
In terms of computational graphs, extrapolation based on atomic mechanistic knowledge corresponds to predicting the immediate next state based on the current observations and intended action. 
Formally, the inference process can be expressed as predicting 
$$S_t \sim P(S_t \mid S_{t-1}, a_{t-1}, \ldots, S_{t-h}, a_{t-h}),$$
where $h$ denotes the window size of historical states. All of $S_{t-1} \ldots S_{t-h}$ are observed.

\paragraph{Transitive Prediction.}
A WM that only predicts the immediate next state is hardly useful for complex planning in long-horizon tasks.
Given a long hypothetical action sequence generated from any policy, a competent WM should accurately predict the corresponding future state.
The statistically less biased way is to perform a step-by-step extrapolation into distant future states \cite{prystawski2023thinkstepstepreasoning}. 
Known as transitive inference, this ability is exhibited by many intelligent animals, including rats, monkeys, and human infants \cite{Wright17112015, thompson2007statistical, Mannella_2024, Bryant1971TransitiveIA, mcgonigle1977monkeys}.
Formally, the inference process can be expressed as:
\begin{align*}
(\widehat{S_{t+q}}, \widehat{S_{t+q-1}}, &\ldots, \widehat{S_t}) \\
\sim P_\theta(&\widehat{S_{t+q}}, \widehat{S_{t+q-1}}, \ldots, \widehat{S_t} \mid \\
&a_t, \ldots, a_{t+q-1}, a_{t+q}, \\
&S_{t-1}, a_{t-1}, \ldots, S_{t-h}, a_{t-h}),
\end{align*}
and the most natural chain of thoughts (CoT) paradigm can be expressed recursively with: 
\begin{align*}
\widehat{S_{t+i}} \sim 
P_\theta(&\widehat{S_{t+i}} \mid a_{t+i}, \widehat{S_{t+i-1}}, a_{t+i-1}, \ldots, \\
&\widehat{S_t}, a_t, S_{t-1}, a_{t-1}, \ldots, S_{t-h}, a_{t-h}).
\end{align*}

\noindent
All $\widehat{S_i}$ are inferred, not observed. 
In terms of computational graphs, the transitive extrapolation through time takes the form of a forward chain.

\paragraph{Compositional Prediction.} 
Previous research in cognitive psychology provides strong evidence that humans and intelligent animals can adjust their statistical expectations of outcome distributions when domain-specific mechanisms (e.g., agentic preferences, intuitive physics, or sampling procedures) conflict with the base rates of the population \cite{XU200997, Eckert2021TheAL,  denison2014probability, doi:10.1073/pnas.1003095107, teglas2011pure}.
Extending from sampling processes to general next-state predictions, we identify another advanced inference mechanism: \textbf{integrating two or more known conflicting or synergistic mechanisms into a unified effect}.

We provide a motivating example to demonstrate what we mean by compositional inference:
Consider a 2D plane with three balls, A, B, and C, each of equal weight.
\begin{itemize}[leftmargin=10pt, topsep=1pt, noitemsep]
    \item \textbf{Observation 1}: Ball A strikes Ball C from the lower left at a specific speed and angle, causing C to move upper right.
    \item \textbf{Observation 2}: Ball B strikes Ball C from the vertically symmetric lower right at the same speed, causing C to move upper left.
\end{itemize}
Now, suppose the agent has never observed a scenario where a single ball is simultaneously struck by two others. 
However, with basic physical intuition, the agent should infer that the leftward and rightward motion components cancel each other out, leading to a prediction that Ball C will move vertically upward.

Formally expressing compositional inference under the standard Markov Decision Process will be tricky \cite{van2012reinforcement}, because the monolithic state denotation $S$ in the MDP formalism fails to capture a crucial fact that complex states can be decomposed into multiple atomic states.
To bridge the gap, we define a conceptual-level composition function \cite{marr2010vision} $S_t = \textrm{Compose}\big[S_{t}^{(1)}, \ldots S_t^{(n)}\big]$ to denote the whole relationship between $n$ component states and complex states.
Then compositional inference can be expressed as:
$$ S_t \sim \textrm{Compose}\big[S_t^{(1)}, \ldots S_t^{(n)})$$
where $S_t^{(i)} \sim P_\theta(S_t^{(i)} \mid S_{t-1}^{(i)}, a_{t-1}, \ldots S_{t-h}^{(i)}, a_{t-h}\big]$.
The compositional extrapolation through time takes the form of a collider with two or more parent node.

\subsection{Final Remarks}
In this work and the conceptual framework section, we adopt a specific interpretation of a world model, which we refer to as a \textit{mechanistic world model}.
For example, a statistical model (e.g., multi-linear regression, XGBoost) used to predict a client's risk of loan default based on features like yearly income, number of children, ethnicity, and medical history is undeniably a form of future-state prediction. 
However, such a model does not faithfully simulate world dynamics, as it lacks any representation of temporal progression. 
While it may provide useful predictive insights, it does not qualify as a mechanistic model because it blackboxes causal mechanisms and interactive kinetic dynamics.
Thus, it is important to recognize that mechanistic simulation is not the only approach to extrapolating future states.

%% file: Appendices/a2_benchmark.tex
\setcounter{tocdepth}{2}
\section{Benchmark Details and Comparisons}

\subsection{Benchmark Statistics}
\label{appendix:benchmark_breakdown}
Table \ref{tab:benchmark} provides a structured overview of our benchmark tasks across different perceptual and predictive dimensions.
\input{Appendices/task_description}
\input{tables/task_description}

\input{tables/bench_compare}

\subsection{Relevant Benchmarks}
\label{sec:related_bench}

We summarize and compare \benchmark to existing benchmarks in Table~\ref{tab:benchmarks}, and provide more descriptive details below.

\paragraph{Visual semantics perception benchmark.}
This line of work primarily tests multimodal models’ static (image) and dynamic (video) recognition competency of object class, object existence, object components, object properties, postures, actions, agent-object interactions, activities, emotions, social relations, functions, image quality, image style, scene, OCR, layouts. Such competency requires models to have diverse schematic knowledge and common sense about the anthropocentric world, and strong pattern recognition capability (i.e., recognize unobserved instances of known categories). However, the foundational perceptual and cognitive functions as enumerated in our work are out of coverage. Representative works in this line include MMBench\cite{liu2024mmbench}, MvBench \cite{li2024mvbench},  VSR\cite{liu2023visual}, VQA v2 \cite{Goyal_2018}, UOUO \cite{pi2024uouo}, MME \cite{fu2024mmecomprehensiveevaluationbenchmark}, STAR \cite{wu2024star}, Perception Test \cite{pătrăucean2023perception}, MSRVTT-QA \cite{Chen_2022}, NExT-QA \cite{xiao2021nextqanextphasequestionansweringexplaining}.

\paragraph{Visual perception Benchmark.}
This line of work primarily focuses on testing models’ competency of low-level, semantic-scarce visual perceptions, such as elementary visual attributes recognition (e.g. color, material, shape, size, texture), line segments, lighting, optical flow, insection, segmentation, overlapping area, corresponding points across different perspectives.
CLEVR \cite{johnson2016clevr}, BLINK \cite{fu2024blink}, BlindTest \cite{rahmanzadehgervi2024visionlanguagemodelsblind}, V* \cite{wu2023vguidedvisualsearch}, KITTI  \cite{geiger2012we}, ActiView \cite{wang2024activiewevaluatingactiveperception}, Shitsukan-eval \cite{shitsukan}. 

\paragraph{Spatiotemporal and motion perception Benchmark.}
This spatial perception tasks focus on evaluating models' ability to understand spatial relationships, configurations, and arrangements within static and dynamic contexts. Spatial perception tasks require recognizing geometric and topological relationships between objects, including proximity, alignment, containment, intersection, adjacency, relative positions, orientation, and distances, as well as multi-perspective alignment and integration. Rel3D \cite{goyal2020rel3dminimallycontrastivebenchmark}, SPARTQA \cite{mirzaee-etal-2021-spartqa}, SpatialSense \cite{yang2019spatialsenseadversariallycrowdsourcedbenchmark}, 3D-Shape-Test \cite{eppel2024largelanguagevisionmodels}, DriveMLLM \cite{guo2024drivemllmbenchmarkspatialunderstanding}, VSI-Bench \cite{yang2024thinkingspacemultimodallarge}.

Temporal and motion perception go hand-in-hand. Since perception of time typically relies on changes and motions, temporal and motion perception tasks lack clear boundaries. Temporal perceptions typically involve: action count, attribute change, action sequence and procedure understanding, event order, scene transition, character order, and action antonym. Motion perception typically involves: direction (e.g. left, clockwise, up, outward), speed, and trajectory. Representative works include: TOMATO \cite{shangguan2024tomatoassessingvisualtemporal}, CATER \cite{girdhar2020caterdiagnosticdatasetcompositional}.

\paragraph{Next state prediction benchmark.}
Next-state prediction is different from the perception, visual-semantic inference (typically about static, state-intrinsic specifications, such as categories, properties, functions, relations, emotions, and intentions), and verbal-logical reasoning (e.g. comparison, logic operations) tasks introduced above. Emphasizing extrapolating objective world states, next-state prediction requires grounded knowledge (in contrast with verbal knowledge) of world mechanics and dynamics. That is, how the environment changes and transits. The agent-centric next-state prediction would additionally emphasize action-transition and interactive dynamics. Representative works includes: Physion \cite{bear2022physion}, Physion++ \cite{tung2023physionevaluatingphysicalscene}, Phyre \cite{bakhtin2019phyrenewbenchmarkphysical}, CLEVRER \cite{yi2020clevrercollisioneventsvideo}, IntPhys \cite{riochet2020intphysframeworkbenchmarkvisual}, CoPhy \cite{baradel2020cophycounterfactuallearningphysical}, CRIPP-VQA\cite{patel2022crippvqacounterfactualreasoningimplicit},
 SEED-Bench \cite{li2023seedbenchbenchmarkingmultimodalllms}.

%% file: Appendices/task_description.tex
\subsubsection{Perception}

\paragraph{Spatial Perception}
\begin{enumerate}[leftmargin=10pt, topsep=1pt, noitemsep]
    \item Spatial Relation (SR): Given the front and top view of two objects on a table, let the model infer the relative position of one object with respect to the other. This task evaluates whether the model can accurately discern spatial relationships based on visual cues.
    \item Spatial Vacancy (SE-V): Given an object and a structure with a hollow space in the middle, let the model infer whether the object can fit into the hollow space. Thus testing the model’s understanding of spatial constraints.
    \item Spatial Occupancy (SE-O): Given two objects with different sizes, let the model infer which object is larger. Thus testing the model's understanding of object size.
    \item Spatial Positioning (SP): Given the front and side view of a set of object arrangements, let the model infer the top view of the objects. 
\end{enumerate}
These multi-view tasks emphasize the model’s ability to synthesize distinct viewpoints into a coherent three-dimensional representation of object arrangements.

\paragraph{Temporal Perception}
\begin{enumerate}[leftmargin=10pt, topsep=1pt, noitemsep]
    \item Temporal positioning (TP): Compare two episodes of object motions from different views, and let the model differentiate which motion started first.
    \item Temporal extension (TE): Compare two episodes of object motions from different perspectives, and let the model differentiate which motion lasts longer.
\end{enumerate}
Collectively, these tasks investigate the aptitude of a model to maintain consistent temporal representations, estimate durations, and infer the correct order of events.

\paragraph{Visual Perception}
\begin{enumerate}[leftmargin=10pt, topsep=1pt, noitemsep]
    \item Color (V-C): Differentiating whether two objects have the same color, or identifying which color the object is.
    \item Shape (V-S): Determine the object's shape; another task involves differentiating whether two objects have the same shape.
    \item Material (V-M): Differentiating whether the two objects have the same material or the model determine the material of the given object. 
\end{enumerate}
By isolating these fundamental visual features, the tasks provide targeted evaluations of how effectively a model can parse and differentiate basic object attributes, separate from any contextual or motion-based confounding factors.

\paragraph{Motion Perception}
\begin{enumerate}[leftmargin=10pt, topsep=1pt, noitemsep]
    \item Motion Identification (MI): Given an episodes of motions of an object and a set of static objects, let the model decide which object is moving. This setup tests the model's capacity to identify the moving action of objects.
    \item Motion Speed (MS): Given episodes of motions of two objects, let the model decide which object moves faster. This setup tests the model’s capacity to track position changes over time and estimate relative velocity.
    \item Motion Direction (MD): Given one moving object and a set of static objects, let the model determine which static object the moving object is heading towards. This setup tests the model’s capacity to track position changes over time and estimate relative moving direction.
    \item Motion Trajectory (MT): Given an episode of motions of two objects, let the model decide the moving trajectory of a specified object, assessing its aptitude for higher-level spatiotemporal pattern recognition and object-specific path tracking across multiple frames.
\end{enumerate}

\paragraph{Quantitative Perception}
\begin{enumerate}[leftmargin=10pt, topsep=1pt, noitemsep]
    \item Discrete Quantity (DQ): Given the top view of objects on the table, let the model count the objects. This setup evaluates the model capacity for discrete numerical estimation.
    \item Continuous Quantity (CQ): Given the top view of two objects with the same color theme, let the model determine which object has a darker shade.
    \item Relative Quantity (RQ): Given the top view of objects with different colors, determine which color group has more objects. This setup probes counting skills, numerical reasoning, and perceptual comparisons in a visual context.
\end{enumerate}

\subsubsection{Prediction}
\paragraph{Mechanistic Knowledge}
\begin{enumerate}[leftmargin=10pt, topsep=1pt, noitemsep]
    \item Intuitive Physics (M-IP): Given a sequence of images showing consecutive states of the environment in which two objects move towards each other, let the model choose the most probable prediction of the next state. This setup evaluates the model capacity in physical reasoning.
    \item Agent Navigation (M-Nav): Given an image of the start state, let the model choose what is most likely to be the final state after the robot/vehicle moves in a certain direction. This setup evaluates the model capacity in predictive reasoning.
    \item Agent Manipulation (M-Man): Given an image of the start state, let the model choose what is most likely to be the final state after the robot arm does certain movements toward the object. This setup evaluates the model capacity in predictive reasoning for robot manipulation.
\end{enumerate}

\paragraph{Transitivity}
\begin{enumerate}[leftmargin=10pt, topsep=1pt, noitemsep]
    \item Agent Navigation (T-Nav): Given an image of the start state, let the model choose what is most likely to be the final state after the robot/vehicle moves through multiple directions in sequence. This setup evaluates the model capacity in predictive reasoning for autonomous navigation.
    \item Agent Manipulation (T-Man): Given an image of the start state, let the model choose what is most likely to be the final state after the robot arm performs two actions in sequence. This setup evaluates the model capacity in multi-step predictive reasoning for robotic manipulation.
\end{enumerate}

\paragraph{Compositionality}
\begin{enumerate}[leftmargin=10pt, topsep=1pt, noitemsep]
    \item Multi-Object Intuitive Physics (C-IP): Given several images of two balls colliding with a third object at the same time, let the model predict the state after the collision occurred. This task evaluates the model’s ability to perform compositional inferences about physical causality and object behavior.
    \item Multi-Agent Manipulation (C-Man): Given an image of the start state, let the model choose what is most likely to be the final state after two robot arms do certain actions on one object simultaneously. This setup evaluates the model capacity in concurrent action predictive reasoning for robotic manipulation.
\end{enumerate}

%% file: tables/task_description.tex
\newcolumntype{C}[1]{>{\centering\arraybackslash}m{#1}}
\begin{table*}[ht!]
\centering
\scriptsize
\renewcommand{\arraystretch}{1.5}
\setlength{\extrarowheight}{2.5pt}
\begin{tabular}{|p{2.65cm}|C{4.5cm}|C{4.25cm}|C{4.0cm}|}
\hline
\textbf{Dimension} & \textbf{Subdimension} & \textbf{Total Cases} & \textbf{\# Images per Case} \\
\hline
\multirow{4}{*}[-2em]{ \centering
 \makecell{
   \textbf{Spatial Perception}
 }}
& Spatial Relation (SR) 
&\makecell{1600 (ManiSkill) \\ 8640 (ThreeDWorld)}
& \makecell{2 (ManiSkill) \\ 2 (ThreeDWorld)}\\
\cline{2-4}
& Spatial Vacancy (SE-V) 
& \makecell{2880 (ManiSkill) \\ 324 (ThreeDWorld)}
& \makecell{3 (ManiSkill) \\ 1 (ThreeDWorld)}\\
\cline{2-4}
& Spatial Occupancy (SE-O) 
& \makecell{1152 (ManiSkill) \\ 941 (ThreeDWorld)}
& \makecell{1 (ManiSkill) \\ 1 (ThreeDWorld)}\\
\cline{2-4}
& Spatial Positioning (SP) 
& \makecell{6400 (ManiSkill)\\2000 (ThreeDWorld)}
& \makecell{2 (ManiSkill)\\2 (ThreeDWorld)}\\
\hline

\multirow{2}{*}[-1em]{\centering
 \makecell{
   \textbf{Temporal Perception}
 }}
& Temporal Positioning (TP) 
& \makecell{2000 (ManiSkill) \\ 5646 (ThreeDWorld)}
& \makecell{5 (ManiSkill) \\6 (ThreeDWorld)}\\
\cline{2-4}
& Temporal Extension (TE) 
& \makecell{ 2000 (ManiSkill) \\ 7680 (ThreeDWorld)}
& \makecell{5 (ManiSkill) \\ 6 (ThreeDWorld)}\\
\hline

\multirow{3}{*}[-0.3em]{\centering
 \makecell{
   \textbf{Visual Perception}
 }}
& Color (V-C) 
& \makecell{ 648 (ManiSkill) \\1440 (ThreeDWorld)}
& \makecell{3 (ManiSkill) \\3 (ThreeDWorld)}\\
\cline{2-4}
& Shape (V-S) 
& \makecell{ 1000 (ThreeDWorld)}
& \makecell{3 (ThreeDWorld)}\\
\cline{2-4}
& Material (V-M) 
& \makecell{1728 (ThreeDWorld)}
& \makecell{1 (ThreeDWorld)}\\
\hline

\multirow{4}{*}[-2em]{\centering
 \makecell{
   \textbf{Motion Perception}
 }}
& Motion Identification (MI) 
& \makecell{2688 (ManiSkill)\\1913 (ThreeDWorld)}
& \makecell{6 (ManiSkill)\\4 (ThreeDWorld)}\\
\cline{2-4}
& Motion Speed (MS) 
& \makecell{2304 (ManiSkill)\\ 2304 (ThreeDWorld)}
& \makecell{4 (ManiSkill)\\ 8 (ThreeDWorld)}\\
\cline{2-4}
& Motion Direction (MD) 
& \makecell{2688 (ManiSkill)\\1913 (ThreeDWorld)}
& \makecell{6 (ManiSkill)\\6 (ThreeDWorld)}\\
\cline{2-4}
& Motion Trajectory (MT) 
& \makecell{6912 (ManiSkill)\\671 (ThreeDWorld)}
& \makecell{7 (ManiSkill)\\8 (ThreeDWorld)}\\
\hline

\multirow{3}{*}[-2em]{\centering
 \makecell{
   \textbf{Quantitative Perception}
 }}
& Discrete Quantity (DQ) 
& \makecell{2520 (ManiSkill)\\1134 (ThreeDWorld)}
& \makecell{1 (ManiSkill)\\1 (ThreeDWorld)}\\
\cline{2-4}
& Continuous Quantity (CQ) 
& \makecell{756 (ThreeDWorld)}
& \makecell{1 (ThreeDWorld)}\\
\cline{2-4}
& Relative Quantity (RQ) 
& \makecell{2880 (ManiSkill)\\1701 (ThreeDWorld)} & \makecell{1 (ManiSkill)\\1 (ThreeDWorld)}\\
\hline

\multirow{3}{*}[-2em]{\centering
 \makecell{
   \textbf{Mechanistic Knowledge}\\
   \textbf{(Prediction)}
 }}
& Intuitive Physics (M-IP) 
& \makecell{ 2160 (Physion: Slide) \\3456 (Physion: Drop) \\1728(Physion: Collide)}
& \makecell{ 3 (Physion: Slide) \\3 (Physion: Drop) \\3 (Physion: Collide)}\\

\cline{2-4}
& Agent Navigation (M-Nav)
& \makecell{\\1575 (Carla) \\ 1728 (Habitat-Lab)\\}
& \makecell{\\1 (Carla)\\ 1 (Habitat-Lab)\\}\\
\cline{2-4}
& Agent Manipulation (M-Man)
& \makecell{\\2304 (ManiSkill: Drop) \\2304 (ManiSkill: Lift) \\2304 (ManiSkill: Push)}
& \makecell{1 (ManiSkill: Drop) \\1 (ManiSkill: Lift) \\1 (ManiSkill: Push)}\\
\hline

\multirow{2}{*}[-2em]{\centering
 \makecell{
   \textbf{Transitivity (Prediction)}
 }}
& Agent Navigation (T-Nav) 
& \makecell{700 (Carla) \\ 1956 (Habitat-lab)}
& \makecell{\\1 (Carla) \\ 1 (Habitat-lab)}\\
\cline{2-4}
& Agent Manipulation (T-Man) 
& \makecell{\\1728 (ManiSkill: Push\&Pick)\\ 576 (Maniskill: Pick\&Rotate)}
& \makecell{\\1 (ManiSkill: Push\&Pick\\ Pick\&Rotate)}\\
\hline

\multirow{2}{*}[-0.8em]{\centering
 \makecell{
   \textbf{Compositionality}\\
   \textbf{(Prediction)}
 }}
& Multi-Object Intuitive Physics (C-IP) 
& \makecell{1296 (ThreeDWorld)}
& \makecell{\\3 (ThreeDWorld)}\\
\cline{2-4}
& Multi-Agent Manipulation (C-Man) 
& \makecell{\\2304 (ManiSkill: Push)\\2304 (ManiSkill: Lift)}
& \makecell{\\1 (ManiSkill: Push)\\1 (ManiSkill: Lift)}\\
\hline
\end{tabular}
\caption{Benchmark Task Breakdown}
\label{tab:benchmark}
\end{table*}

%% file: tables/bench_compare.tex
\begin{table*}[t]
\centering
\scalebox{0.75}{
\begingroup
\renewcommand{\arraystretch}{0.8}
\setlength{\tabcolsep}{1.1pt}
\hspace{-10pt}
\begin{tabular}{cccccccccccccccccccccccc}
\toprule
\multirow{2}{*}{Benchmark} & \multicolumn{4}{c}{Spatial}                                                                                   & \multicolumn{2}{c}{Temporal}                    & \multicolumn{3}{c}{Quantity}                                                           & \multicolumn{3}{c}{Visual}                                                                    & \multicolumn{4}{c}{Motion}                                                                                                    & \multicolumn{3}{c}{Mechanistic}                                                         & \multicolumn{2}{c}{Transitivity}                        & \multicolumn{2}{c}{Compositionality}              \\
\cmidrule(lr){2-5} \cmidrule(lr){6-7} \cmidrule(lr){8-10}
\cmidrule(lr){6-7} \cmidrule(lr){11-13} \cmidrule(lr){14-17} \cmidrule(lr){18-20} \cmidrule(lr){21-22} \cmidrule(lr){23-24}
                           & SP                   & MV        & SO        & SE & TP & TE & DQ        & CQ        & RQ & AR-S      & AR-M      & AR-C      & MD        & DI        & SC        & MT        & IP        & Nav & Mani      & Nav & Mani      & IP & Mani \\
\cmidrule(lr){1-1} \cmidrule(lr){2-5} \cmidrule(lr){6-7} \cmidrule(lr){8-10}
\cmidrule(lr){6-7} \cmidrule(lr){11-13} \cmidrule(lr){14-17} \cmidrule(lr){18-20} \cmidrule(lr){21-22} \cmidrule(lr){23-24}
BLINK \shortcite{fu2024blink}                     & \ding{52}            & \ding{52} &                               &                        &                        &                        & \ding{52} & \ding{52} &                        &                               &                               &                               &                               & \ding{52} &                               &                               &                               &                         &                               &                         &                               &                        &                          \\
SpatialRGPT \shortcite{cheng2024spatialrgpt}         & \ding{52}            &                               & \ding{52} &                        &                        &                        &                               &                               &                        &                               &                               &                               &                               &                               &                               &                               &                               &                         &                               &                         &                               &                        &                          \\
VSI-Bench \shortcite{yang2024thinkingspacemultimodallarge}                 & \ding{52}            &                               & \ding{52} &                        &                        &                        & \ding{52} &                               &                        &                               &                               &                               &                               &                               &                               &                               &                               &                         &                               &                         &                               &                        &                          \\
VL-CheckList  \shortcite{zhao2023vlchecklistevaluatingpretrainedvisionlanguage}             & \ding{52}            &                               & \ding{52} &                        &                        &                        &                               &                               &                        &                               & \ding{52} & \ding{52} &                               &                               &                               &                               &                               &                         &                               &                         &                               &                        &                          \\
CLEVR \shortcite{johnson2016clevr}                     & \ding{52}            &                               & \ding{52} &                        &                        &                        & \ding{52} &                               &                        & \ding{52} & \ding{52} & \ding{52} &                               &                               &                               &                               &                               &                         &                               &                         &                               &                        &                          \\
CLEVRER \shortcite{yi2020clevrer}                   & \ding{52}            &                               & \ding{52} &                        &                        &                        & \ding{52} &                               &                        & \ding{52} & \ding{52} & \ding{52} &                               &                               &                               &                               & \ding{52} &                         &                               &                         &                               &               \ding{52}         &                          \\
NLVR \shortcite{suhr-etal-2017-corpus}                      & \ding{52}            &                               & \ding{52} &                        &                        &                        &                               &                               &                        & \ding{52} &                               & \ding{52} &                               &                               &                               &                               &                               &                         &                               &                         &                               &                        &                          \\
NLVR2 \shortcite{suhr-etal-2019-corpus}                     & \ding{52}            &                               &                               &                        &                        &                        & \ding{52} &                               &                        &                               &                               &                               &                               &                               &                               &                               &                               &                         &                               &                         &                               &                        &                          \\
ShapeWorld \shortcite{kuhnle2017shapeworldnewtest}                & \ding{52}            &                               &                               &                        &                        &                        & \ding{52} &                               &                        & \ding{52} &                               & \ding{52} &                               &                               &                               &                               &                               &                         &                               &                         &                               &                        &                          \\
VALSE \shortcite{parcalabescu-etal-2022-valse}                     & \ding{52}            &                               &                               &                        &                        &                        & \ding{52} &                               &                        &                               &                               &                               &                               &                               &                               &                               &                               &                         & \ding{52} &                         &                               &                        &                          \\
MVBench \shortcite{Li_2024_CVPR}                   & \ding{52}            &                               &                               &                        &                        &                        & \ding{52} &                               &                        & \ding{52} &                               & \ding{52} &                               & \ding{52} &                               &                               &                               &                         &                               &                         &                               &                        &                          \\
MMBench \shortcite{10.1007/978-3-031-72658-3_13}                   & \ding{52}            &                               &                               &                        &                        &                        & \ding{52} &                               &                        & \ding{52} & \ding{52} & \ding{52} &                               &                               &                               &                               &                               &                         &                               &                         &                               &                        &                          \\
MME \shortcite{fu2024mmecomprehensiveevaluationbenchmark}                       & \ding{52}            &                               &                               &                        &                        &                        & \ding{52} &                               &                        &                               &                               & \ding{52} &                               &                               &                               &                               &                               &                         &                               &                         &                               &                        &                          \\
VQA(v2) \shortcite{Goyal_2018}                   & \ding{52}            &                               &                               &                        &                        &                        & \ding{52} &                               &                        &                               &                               & \ding{52} &                               &                               &                               &                               &                               &                         &                               &                         &                               &                        &                          \\
NExT-QA \shortcite{xiao2021nextqanextphasequestionansweringexplaining}                   & \ding{52}            &                               &                               &                        &                        &                        & \ding{52} &                               &                        &                               &                               &                               &                               &                               &                               &                               &                               &                         &                               &                         &                               &                        &                          \\
V* \shortcite{wu2023vguidedvisualsearch}                        & \ding{52}            &                               &                               &                        &                        &                        & \ding{52} &                               &                        &                               & \ding{52} & \ding{52} &                               &                               &                               &                               &                               &                         &                               &                         &                               &                        &                          \\
ActiView \shortcite{wang2024activiewevaluatingactiveperception}                   & \ding{52}            &                               &                               &                        &                        &                        & \ding{52} &                               &                        &                               &                               &                               &                               &                               &                               &                               &                               &                         &                               &                         &                               &                        &                          \\
SEED-Bench \shortcite{li2023seedbenchbenchmarkingmultimodalllms}                & \ding{52}            &                               &                               &                        &                        &                        & \ding{52} &                               &                        & \ding{52} & \ding{52} & \ding{52} &                               &                               &                               &                               &                               &                         &                               &                         &                               &                        &                          \\
Perception Test \shortcite{pătrăucean2023perception}           & \ding{52}            &                               &                               &                        &                        &                        &                               &                               &                        & \ding{52} &                               &                               & \ding{52} &                               &                               &                               & \ding{52} &                         &                               &                         &                               &               \ding{52}         &                          \\
BlindTest \shortcite{rahmanzadehgervi2024visionlanguagemodelsblind}                 &                          &                               &                               &                        &                        &                        & \ding{52} &                               &                        &                               &                               & \ding{52} &                               &                               &                               &                               &                               &                         &                               &                         &                               &                        &                          \\
SHAPES \shortcite{neuralmodulenetwork2016}                   &                          &                               &                               &                        &                        &                        &                               &                               &                        & \ding{52} &                               & \ding{52} &                               &                               &                               &                               &                               &                         &                               &                         &                               &                        &                          \\
TOMATO \shortcite{shangguan2024tomatoassessingvisualtemporal}                    &                          &                               &                               &                        &                        &                        &                               &                               &                        & \ding{52} &                               &                               &                               & \ding{52} & \ding{52} & \ding{52} &                               &                         &                               &                         &                               &                        &                          \\
STAR \shortcite{wu2024star}                      &                          &                               &                               &                        &                        &                        &                               &                               &                        &                               &                               &                               &                               &                               &                               &                               &                               &                         & \ding{52} &                         & \ding{52} &                        &                         \\
\cmidrule(lr){1-1} \cmidrule(lr){2-5} \cmidrule(lr){6-7} \cmidrule(lr){8-10}
\cmidrule(lr){6-7} \cmidrule(lr){11-13} \cmidrule(lr){14-17} \cmidrule(lr){18-20} \cmidrule(lr){21-22} \cmidrule(lr){23-24}
\textbf{Ours}                       & \ding{52}            & \ding{52} & \ding{52} & \ding{52} & \ding{52} & \ding{52} & \ding{52} & \ding{52} & \ding{52} & \ding{52} & \ding{52} & \ding{52} & \ding{52} & \ding{52} & \ding{52} & \ding{52} & \ding{52} & \ding{52} & \ding{52} & \ding{52} & \ding{52} & \ding{52} & \ding{52}   \\
\bottomrule
\end{tabular}
\endgroup}
\vspace*{-5pt}
\caption{Comparison between different benchmarks.}
\vspace*{-15pt}
\label{tab:benchmarks}
\end{table*}

%% file: Appendices/a3_simulatoir.tex
\section{Simulator Setup}
\label{appendix:sim_setup}

We describe how we set up each simulator for generating test cases for the proposed world model benchmark.

\subsection{ThreeDWorld}
We use the ThreeDWorld simulator \citep{threedworld} to generate images for the perception tasks, with intuitive physics (e.g., collisions) modeled using the Physion framework \citep{bear2022physion}.
We first select a curated set of pre-packaged scenes, objects, and materials from ThreeDWorld. 
For most questions, we spawn selected objects like cubes and spheres onto the floor or a table within one of the selected scenes, then randomly varying their color, size, position, and material to ensure diversity across samples. 
We then render images of the scene from multiple viewing angles, including top-down, front, and side views, to ensure diverse perspectives for question generation.
For tasks requiring a sequence of images to demonstrate object movement, we first render an initial image capturing the objects in their original spawned positions. 
We then teleport the objects to new locations, rendering an image after each movement until the desired movement is complete.
After generating all images for a given scene, we clear the scene by removing all objects. We then repeat the process in a newly selected scene, beginning with the random selection of object color, size, position, and material to ensure diversity across scenes.

\subsection{ManiSkill}

We use both ManiSkill framework version 2 \cite{gu2023maniskill2} and 3 \cite{taomaniskill3}) and we do the following to generate the images used in the dataset.
For most questions, we spawn objects such as cubes and spheres onto a table in the ReplicaCAD apartment scene \citep{szot2021habitat} and render an image of the scene.
For tasks requiring before-and-after images, we first render the scene, then teleport objects to new locations, and render again.
For static tasks, following a similar approach to ThreeDWorld, we place several cubes and spheres on a table within the provided background. 
To ensure sample diversity, these objects vary randomly in color, size, and position. 
We then render images from multiple viewpoints (e.g., top-down, front, and side angles) to provide diverse perspectives for subsequent question generation
For tasks requiring a sequence of images to capture object movements (similar to ThreeDWorld), we first render an initial image depicting the objects in their original positions. We then relocate the objects, rendering an image after each repositioning, until the intended motion is complete.
For agent manipulation questions, we use the RoboCasa dataset \citep{robocasa2024}. 
One scene features two Franka Panda arm robots, while another includes only one.
Using ManiSkill, we randomize several dimensions in our test cases, including object geometry, object color, apartment layout, and visual style. 
Teleoperation tools collect demonstrations of both successful and failed trajectories for each object geometry, with physics simulation enabled. Other randomized dimensions are synthetically generated.
For the question-answer pairs, we render the first and last frames of each demonstration

\subsection{Physion}

We use various setups (e.g. collide, slide, drop) provided by Physion~\citep{bear2022physion} and instantiate various objects and run the physics simulation to simulate various physical effects such as dropping or rolling, testing mechanistic state transition knowledge. 
An image is captured before the simulation starts and frames are iteratively being captured after a simulation has run.

\subsection{Carla}

Using Carla \citep{carla} simulator we do the following to generate the images used in the dataset.
We first select a curated set of pre-packaged towns, weather, and car agents from Carla. 
For the majority questions we spawn a car agent at a random position onto one of the selected scenes, instantiated with a random weather from the selected weathers.
We refined the control mechanisms of the car agent to enhance realism, ensuring that actions such as moving forward and turning exhibit natural and physically plausible behavior that aligns with their corresponding natural language descriptions. 
We then provide commands instructing the car to move in a specified direction or make a turn at an intersection. The simulator subsequently renders and captures a sequence of images depicting the car's actions, which are used to construct our dataset.

\subsection{Habitat}

We use Habitat 2 \cite{habitat2} to render the HSSD \cite{khanna2023habitatsyntheticscenesdataset} dataset, which includes a large number of simulated indoor scenes, to create navigational transition action-state pairs. 
We use discrete actions that enable the agent in the simulation to move around and change viewing directions. 
Pre-condition and post-condition images are generated for the dataset.

%% file: Appendices/a4_results.tex
\section{Addendum to Results}
\label{appendix:additional_evidence}

\begin{figure*}[h]
\centering
\includegraphics[width=1.0\textwidth]{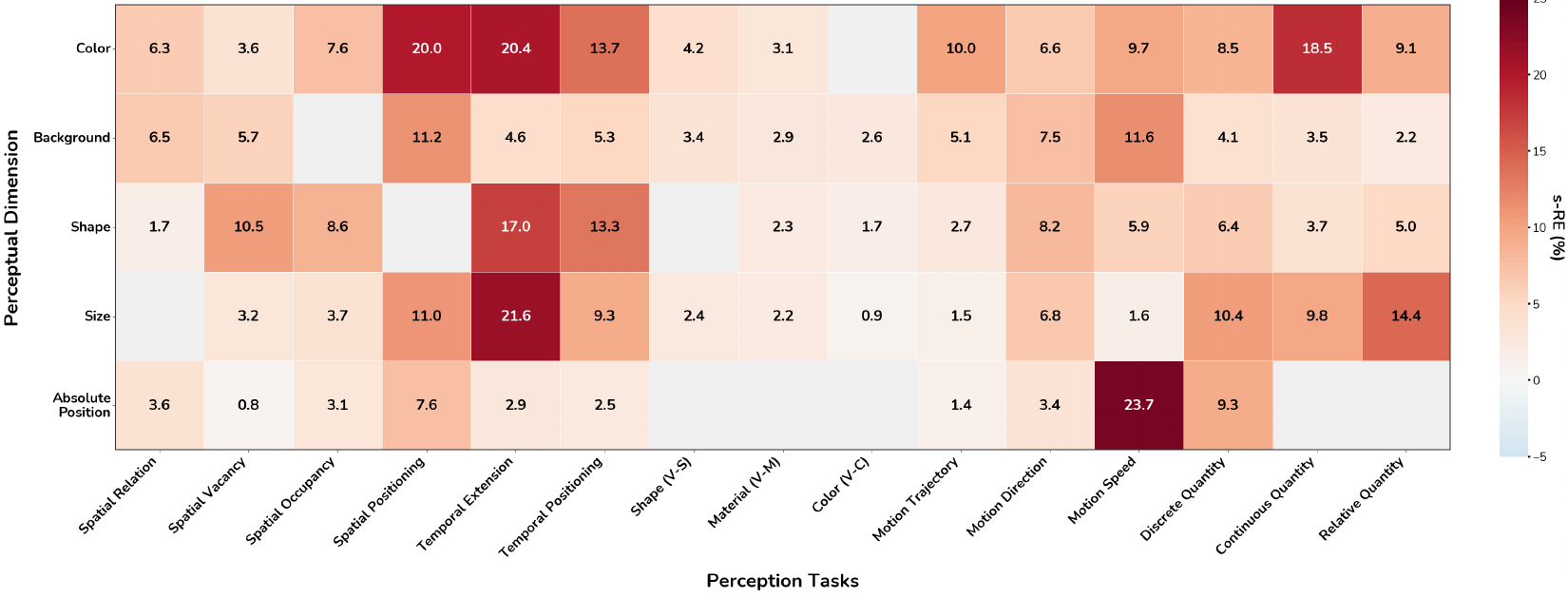}
\caption{Heatmap showing s-RE scores that quantify the relative impact of different physical dimensions on perception task performance. Higher values (darker red) indicate stronger entanglement between a physical dimension and task performance, while lower values (lighter colors) suggest weaker relationships.}
\label{fig:heatmap}
\end{figure*}

\subsection{Entanglement in Perception Tasks }
Figure \ref{fig:heatmap} provides a heatmap showing Relative Entanglement (s-RE) scores. These scores represent the average performance deviation (s-RE) across a subset of the highest-performing models (GPT-4o, Gemini-1.5 Pro, Qwen2-VL, Qwen2.5-VL, and InternVL-2.5).

\subsection{Evaluation via Perceptual Queries}\
\label{appendix:perceptual queris}
We devised three targeted perceptual queries to rigorously assess the models' understanding of dynamic scene attributes:

\begin{itemize} [leftmargin=10pt, topsep=1pt, noitemsep]
    \item Does the scene contain a moving object? 
    \item What is the observed color of the moving object? 
    \item What is the observed shape of the moving object? 
\end{itemize}

\noindent
These queries are intended to verify that the models accurately detect and interpret the moving object—a critical prerequisite for successful task performance. Notably, this stringent evaluation protocol resulted in the exclusion of approximately 30\% of the test instances, thereby underscoring the robustness and effectiveness of our filtering approach.

%% file: Appendices/a5_reproduce.tex
\section{Evaluation and Reproducibility}

\subsection{Human Evaluation}
\label{appendix:human_eval}

We recruited Mechanical Turk Masters on Amazon Mechanical Turk. 
Annotators were required to have a 98\% HIT approval rate, at least 100 approved HITs, and reside in the United States. 
Each problem was evaluated by three annotators, with the final label determined by majority vote (ties were resolved by randomly selecting an answer). 
Workers were paid \$1 per HIT (10 examples per HIT, where each example took about 20 to 30 seconds).
For each task, we provide brief task instructions. Here is an example:

\begin{tcolorbox}[colframe=gray!75!black, boxsep=1mm, left=2mm, right=2mm, top=1mm, bottom=1mm] 
You will be provided with six images, each representing evenly spaced frames from a video. Two moving objects are visible in the frames. Your task is to determine which object started moving first, \texttt{\${object\_name1}} or \texttt{\${object\_name2}}?
\end{tcolorbox}

\subsection{VLM Evaluation}
\label{appendix:prompting_framework}

We develop a general prompt framework to support both open-source and closed-source models under a unified design. The system is built around a general evaluator class capable of loading and executing multiple model types. 
Our data management strategy involves categorizing datasets and assigning each task a unique identifier. This allows seamless retrieval of the corresponding datasets for varied evaluation scenarios. To streamline prompt creation and ensure consistency, we maintain prompt template files containing different formats for a wide range of question types.
Below is the system prompt that we use to regularize the output format from the models:

\begin{tcolorbox}[colframe=gray!75!black, boxsep=1mm, left=2mm, right=2mm, top=1mm, bottom=1mm] 
You are a helpful assistant. You will be given a question to answer. If it is a multichoice question, return the index of your choice 1,2,3,4 or A,B,C,D depending on the question, and then followed by any explanation necessary. If it is a yes/no question, clearly answer ``yes'' or ``no'' at first, and then follow with your explanation if needed. If you are asked to choose the images, please note that the last four images of all the given images are your choices. And your answer should be 1 or 2 or 3 or 4 pointing to these last four images.
\end{tcolorbox}

\subsection{Computational Resource}

We use H100 GPUs to run the experiments, with an estimated total runtime of 200 GPU hours for model inference. 

\subsection{License and Research Artifacts}

\paragraph{Simulators.}

Regarding the licenses or terms for the use and distribution of artifacts, our paper utilizes the following simulation environments: ThreeDWorld~\citep{threedworld}, ManiSkill~\citep{taomaniskill3}, Physion~\citep{bear2022physion}, Carla~\citep{carla}, and Habitat 2~\citep{habitat2}. Detailed documentation on these artifacts is provided in Appendix~\ref{appendix:sim_setup}. Their corresponding licenses are listed in table \ref{table::license}.
 
\input{tables/license}

Having reviewed the rights and terms of these licenses, we confirm that we have fully complied with their requirements. Our work will be released under the  MIT License, ensuring no legal issues arise. As discussed in the ethics statement of section~\ref{ethics}, our benchmark is designed to provide a fundamental evaluation of the core world modeling abilities in VLMs, which generally do not involve social aspects or social reasoning. We stipulate that our work should be used strictly for academic purposes. Since our data is collected from simulators, it is fully anonymized, does not contain personally identifiable information, and does not require additional measures to verify the absence of sensitive information relevant to individuals.

%% file: tables/license.tex
\begin{table}[!h]
\centering 
\begin{tabular}{lll}
    \toprule
    Simulators & URL & License   \\ 
    \cmidrule(lr){1-1} \cmidrule(lr){2-3}
    ThreeDWorld (TDW) & \href{https://github.com/threedworld-mit/tdw}{Link} & BSD-2-Clause \\
    ManiSkill & \href{https://github.com/haosulab/ManiSkill}{Link} & Apache v2.0 \\
    Physion & \href{https://github.com/cogtoolslab/physics-benchmarking-neurips2021}{Link} & MIT license \\
    Carla & \href{https://github.com/carla-simulator/carla}{Link} & MIT license \\
    Habitat 2 & \href{https://github.com/facebookresearch/habitat-lab}{Link} & MIT license \\
    \bottomrule
\end{tabular}
\vspace*{0.25cm}
\caption{License information for the simulators used.}
\label{table::license}
\end{table}